\documentclass[letterpaper, 10 pt, conference]{ieeeconf}
\IEEEoverridecommandlockouts
\usepackage{graphics} 
\usepackage{epsfig} 
\usepackage{amsmath} 
\usepackage{amssymb}  
\usepackage{cite}
\usepackage{algorithm}
\usepackage{algpseudocode}
\usepackage{dsfont}
\usepackage{subfig}
\usepackage{url}
\usepackage{siunitx}
\usepackage{autobreak}
\usepackage{balance}
\usepackage{tabularx}
\usepackage{booktabs}
\usepackage{multirow}
\usepackage{wasysym}
\usepackage{gensymb}
\usepackage{textcomp}
\usepackage[locale=US]{siunitx}

\usepackage{tikz}     
\usepackage{xsavebox} 
\usepackage{atbegshi} 

\usepackage{textcmds}

\newcolumntype{?}{!{\vrule width 1pt}}

\usepackage{float}
\newfloat{algorithm}{t}{lop}

\title{\LARGE \bf
	Clustering of Motion Trajectories by a Distance Measure Based on Semantic Features
}

\newcommand{\R}{\mathds{R}}
\newcommand{\N}{\mathds{N}}

\newcommand\sdots{\makebox[1em][c]{.\hfil.\hfil.}}

\author{Christoph Zelch$^{1}$%
	, Jan Peters$^{2}$%
	, Oskar von Stryk$^{1}$
\thanks{$^{1}$Christoph Zelch and Oskar von Stryk are with the Simulation, Systems, Optimization and Robotics Group, Department of Computer Science,
        TU Darmstadt, Hochschulstr. 10, 64289 Darmstadt, Germany
        {\tt\small \{zelch, stryk\}@sim.tu-darmstadt.de}}%
\thanks{$^{2}$Jan Peters is with the Intelligent Autonomous Systems Group, Department of Computer Science,
	TU Darmstadt, Hochschulstr. 10, 64289 Darmstadt, Germany
    {\tt\small jan.peters@tu-darmstadt.de}}%
}

\makeatletter
\newcommand\notsotiny{\@setfontsize\notsotiny\@vipt\@viipt}
\newcommand\notsotinyCustom{\@setfontsize\notsotiny{6.6}{7.7}}
\makeatother

\xsavebox{PageTopWatermark}{%
    \begin{minipage}{\paperwidth}
        \notsotinyCustom
        \centering
        This is the author's version of an article that has been accepted for 2023 IEEE-RAS 22nd International Conference on Humanoid Robots (Humanoids).\\
        Changes were made to this version by the publisher prior to publication.\\
        2023 IEEE INTERNATIONAL CONFERENCE ON HUMANOID ROBOTS. PREPRINT VERSION. ACCEPTED SEPTEMBER, 2023
    \end{minipage}
}
\xsavebox{PageBottomWatermark}{%
    \begin{minipage}{\paperwidth}
        \notsotinyCustom
        \centering
        \parbox{18cm}{\centering
        \textcopyright 2023 IEEE. Personal use of this material is permitted. Permission from IEEE must be obtained for all other uses, in any current or future media, including reprinting/republishing
        this material for advertising or promotional purposes, creating new collective works, for resale or redistribution to servers or lists, or reuse of any copyrighted component of this work in other works.}
    \end{minipage}
}
\AtBeginShipout{
    \AtBeginShipoutUpperLeft{\raisebox{-0.8cm}{\xusebox{PageTopWatermark}}}
}
\AtBeginShipout{
    \AtBeginShipoutUpperLeft{\raisebox{-27.3cm}{\xusebox{PageBottomWatermark}}}
}

\begin{document}
\maketitle

\thispagestyle{empty}
\pagestyle{empty}

\begin{abstract}
    Clustering of motion trajectories is highly relevant for human-robot interactions as it allows the anticipation of human motions, fast reaction to those, as well as the recognition of explicit gestures. Further, it allows automated analysis of recorded motion data.
    Many clustering algorithms for trajectories build upon distance metrics that are based on pointwise Euclidean distances. However, our work indicates that focusing on salient characteristics is often sufficient.
    We present a novel distance measure for motion plans consisting of state and control trajectories that is based on a compressed representation built from their main features.
    This approach allows a flexible choice of feature classes relevant to the respective task.
    The distance measure is used in agglomerative hierarchical clustering.
    We compare our method with the widely used dynamic time warping algorithm on test sets of motion plans for the Furuta pendulum and the Manutec robot arm and on real-world data from a human motion dataset. The proposed method demonstrates slight advantages in clustering and strong advantages in runtime, especially for long trajectories.
\end{abstract}

\section{Introduction}\label{sec:intro}
Clustering of trajectories is an important tool for analysis, recognition and prediction of motions and has been used in humanoid robotics, e.g., to estimate the gait phase \cite{piperakis_unsupervised_2019}, to learn motion primitives from observations \cite{kulic_incremental_2012}, to analyze posture data \cite{basoeki_clustering_2016}, or in the field of human-robot interactions to build a database structure for human motions that allows the robot to respond faster \cite{yamane_synthesizing_2013}, to recognize hand gestures as input commands \cite{maharani_hand_2018}, \cite{okada_incremental_2010}, \cite{lee_automatic_2006} or to recognize and predict human or robotic actions and movements \cite{yao_trajectory_2017}, \cite{luo_unsupervised_2018}, \cite{sung_trajectory_2012}.

Trajectories are typically represented as sequences of data points associated with time stamps. They can be extracted from image data, the output of a motion capture system or internal sensors that provide data for trajectories at discrete time steps. This data format can be interpreted as linear splines. 
Trajectory data may also originate from numerical planning or solutions of optimal control problems (for example, Merkt et al. \cite{merkt_memory_2021}).
Trajectories resulting from the latter can be represented as splines (e.g., in DIRCOL \cite{stryk_numerical_1993}), potentially cubic or even quintic splines to provide differentiability and smoothness.
One application is the clustering of motion plans from different start states provided by the optimal control software DIRCOL \cite{stryk_numerical_1993} to differentiate between distinct solution clusters.
In this work, we refer to motion plans as a combination of a state and a control trajectory. The distance measure is applicable to general trajectories as well as motion plans.

The insight that motivates this work is that the exact path of a trajectory is less relevant for clustering. Much more important is the sequence of some meaningful features (e.g., extrema, changepoints, roots, joint limits) and their salience in the trajectory.
We thus do not focus on measuring small differences between very similar trajectories as this is, e.g., done when the error between a reference and some executed trajectory is computed. Instead, we are interested in a measure that captures differences in the general characteristic shape of trajectories to use this measure to cluster a set of trajectories into groups of similar graphs.

\paragraph*{Contribution}
In this work, we present a novel approach to transform robotic trajectory data into a sequence of high-level semantic features (which are characteristic points) describing the general shape of its graph. These features are supplemented with information about their temporal position and their salience in the trajectory. This feature sequence is used in an existing sub-sequence-based distance metric presented by Elzinga et al. \cite{elzinga_versatile_2012}, which we adapt for our application. The resulting distance measure for trajectories is used in agglomerative hierarchical clustering to identify trajectories with similarly shaped graphs. The advantage of this approach is that relevant feature classes can be chosen depending on the application or the considered problem.
Our trajectory distance measure is compared with dynamic time warping (DTW) \cite{vaughan_comparing_2016}, we show that our approach 
outperforms this widely used measure on several test sets.

\paragraph*{Related Work}\label{subsec:rel_work}
Gaussian Mixture Models have been used by Lee \cite{lee_automatic_2006}, Piperakis et al. \cite{piperakis_unsupervised_2019} or Luo et al. \cite{luo_unsupervised_2018} for clustering trajectories. However, this approach requires the estimation of multiple Gaussian distributions which can be computationally costly and the result is sensitive to initial points.
Research has been done to cluster trajectory data using (deep) neural networks (NN), e.g., using auto-encoders as in \cite{yao_trajectory_2017, olive_deep_2020, piperakis_unsupervised_2019} or specialized structures like self-organizing incremental NNs \cite{okada_incremental_2010}. NNs can provide good performance on large-scale datasets \cite{olive_deep_2020}. However, they require large amounts of training data to generalize well and provide limited or no insight into how the clustering is done.

Widely used clustering algorithms are hierarchical clustering (Basoeki et al. \cite{basoeki_clustering_2016}, Kuli\'c et al. \cite{kulic_incremental_2012}), k-means (Maharani et al. \cite{maharani_hand_2018}) or density-based clustering.
The vast majority of clustering algorithms require a distance function to measure the distance or similarity of two trajectories. The construction of centroids needed for the k-means algorithm is not intuitive in the context of trajectory clustering. An advantage of hierarchical over density-based clustering is the better interpretability of the results, as hierarchies of clusters can be represented in dendrograms. For these reasons, hierarchical clustering is used in this paper. Nevertheless, the proposed distance measure for trajectories can also be used combined with other distance-based clustering algorithms.

A large part of trajectory distance measures can be divided into those based on Lp-norms (well-known are, e.g., DTW and related, \emph{Fr\'echet} distance) and based on edit distance (like \emph{edit distance with real penalty}, \emph{longest common sub-sequence}) \cite{su_survey_2020}. Their properties have been studied extensively \cite{tao_comparative_2021, vagni_comparison_2021}. These distance measures are typically applied to some condensed representation of the trajectories that are computed, e.g., using principal component analysis as in \cite{piperakis_unsupervised_2019}, multiple correspondence analysis \cite{basoeki_clustering_2016} or hidden Markov models (HMM) \cite{okada_incremental_2010, kulic_incremental_2012}. These reductions all are not intuitive to interpret, HMMs additionally require training.

In the context of social sciences, distance measures are often used to compare sequences of categorical elements from a finite alphabet. Distance measures for semantic sequences are reviewed in \cite{studer_what_2016}. In this work, we build upon the work of Elzinga et al. \cite{elzinga_versatile_2012}, who present a measure called SVRspell, which we adapt to our problem. It is based on the number of matching sub-sequences in strings, which correspond to feature sequences of trajectories. The extraction of features from raw trajectories is related to Schmid et al. \cite{schmid_semantic_2009}, who proposed a trajectory compression method for geographical movement data that uses semantic information to transform raw trajectory data. A review of this topic has been presented by Parent et al. \cite{parent_semantic_2013}. However, there are several differences between the trajectory representations that reflect the specific requirements of our application. In particular, there is no underlying map; consequently, the raw
trajectory cannot be reconstructed from our representation, in contrast to \cite{schmid_semantic_2009}.

\section{Description of the Method}\label{sec_method}

Our clustering of a set of trajectories can be roughly divided into three major steps.
In the first step, we reduce each trajectory to a sequence of high-level features augmented with additional information about timing and salience.
In the second step, we use the sequence-based representation of the trajectories computed in step one to measure distances among them.
The distances from the second step form a distance matrix used in the last step as input for standard agglomerative hierarchical clustering (see \cite{murtagh_algorithms_2017}) to identify similar trajectories in an examined set.
These three steps will be detailed in the following subsections.

\subsection{Definitions}
In motion planning of robotic systems, there typically are state and control trajectories to describe the state of the system in combination with the control values required to generate the motion.
In this paper, we call the combination of a state trajectory $s$ and a control trajectory $c$ synchronized in time \emph{motion plan}:
\begin{equation}\label{eq:def_motion_plans}
T = \left\lbrace s:\left[0, t_f\right]\rightarrow R_s\subset\R^n, c:\left[0, t_f\right]\rightarrow R_c\subset\R^m \right\rbrace,
\end{equation}
where $t_f>0$ is the trajectory end time. We call trajectories with $t_f=1$ time-normalized.
In this work, we represent trajectories as mappings instead of the often-used sequence of values connected with time stamps since it is a better representation of the continuous reality than discrete sequences. Moreover, Equation \eqref{eq:def_motion_plans} gives a more general representation, since these sequences can be interpreted as a continuous linear spline function.

Typical for motion planning for robotic systems is the existence of limits on the codomain of the trajectories that are induced for instance by joint limits or control bounds.
In this work, we thus assume the existence of box-constraints on the codomains of the maps $s$ and $c$, i.e., 
\begin{equation}\label{boundsCodomain}
R_s := \prod\nolimits_{i=1}^{n}[l_{s,i}, u_{s,i}]
\end{equation}
for some lower and upper bounds $l_{s}, u_{s}\in\R^n$ and $R_c$ analogously. Nevertheless, our approach is applicable with small changes to problems without bounds.

\subsection{Step 1: Construction of Sequence-Based Representation}\label{sec:representation_construction}
This subsection describes the extraction of features from time-dependent raw trajectory data.
Following the definition of a multidimensional sequence in \cite{furtado_multidimensional_2016}, we define a \emph{feature sequence} as a sequence of elements $ e_1, \dots, e_q$ where each element $e_i$ is a triplet of attributes $e_i=\left(e^\text{cat}_i, e^\text{time}_i, e^\text{val}_i\right)$. 
In this triplet, $e^\text{cat}_i$ denotes the kind of feature this element represents (\emph{feature class}), $e^\text{time}_i$ gives the time stamp of this feature in the normalized trajectory and $e^\text{val}_i \geq 0$ provides information about the importance of the feature in the sequence. Several examples of such attributes will be given later on. For a given trajectory, we compute $m+n$ sequences, one for each dimension in the state and control (cf. Eq. \eqref{eq:def_motion_plans}).

The selection of relevant features is a crucial part of the application of this method and requires careful design. It depends on the one hand on the features that occur in at least some trajectories of the examined set and on the other hand on the relevance of these features for the problem or application underlying the trajectories.
In this paper, we use maxima ($\wedge$) and minima ($\vee$), active box constraints (upper bound $L_u$, lower bound $L_l$) and for one problem roots ($0$) as feature classes. 
These are the dominant characteristics that are apparent in the trajectories' graphs resulting from our problems. Examples of other features are jumps, steep inclines or roots/signs of derivatives. The selection of relevant characteristics for the distance measure is highly problem specific.

Our state trajectories are represented as cubic splines, the control trajectory as linear splines. This is DIRCOL's output format, typical for collocation-based optimal control solvers. 
For different representations, it may be necessary to adapt the feature extraction approach. Trajectory data from sensors or images is typically a sequence of (potentially multidimensional) time-stamped values. This can be interpreted as linear splines, such that our approach for the control trajectories can be used for this data without further adaptation.

To extract features from a trajectory, we need for each feature class a proceeding to find the times $e^\text{time}_i$ at which the feature $e^\text{cat}_i$ occurs. Furthermore, we need to assign a value $e^\text{val}_i$ to each feature representing its salience in the trajectory shape. The following describes our approach to get these features from linear and cubic splines.

\subsubsection{Extrema}
In linear splines, maxima and minima can only occur at knots (times at which the polynomial changes); we thus search for sign changes between the adjacent splines' slopes to find the extrema.
For cubic splines, we search in each polynomial
for extrema by checking if its derivative, a quadratic function, has roots inside the interval on which the spline is defined. 
In $e_i^\text{cat}$, we distinguish between \emph{maximum} ($\wedge$) and \emph{minimum} ($\vee$).
To assign a value $e^\text{val}$ to all maxima and minima, we use a concept called \emph{(topographic) prominence} \cite{llobera_building_2001} as defined in \cite{kirmse_calculating_2017} that is used, e.g., to value the significance of mountain summits. We slightly adapt this definition and normalize the computed prominence by dividing it by $u_s-l_s$ (or $u_c-l_c$). At this point, we use our assumption of bounded codomains. For unbounded co-domains, one could still normalize the prominence using the inverse tangent or any other mapping $[0, \infty)\rightarrow[0, 1)$.
Since prominence is defined for maxima only, we define the prominence of a minimum of a function $f$ at $t_m$ as the prominence of the maximum of the function $(-f)$ at $t_m$. Finally, we discard all extrema with a prominence lower than a certain threshold to avoid long feature sequences with many irrelevant features.

\subsubsection{Active box constraints}
\begin{figure}
    \centering
    \begin{minipage}{0.235\textwidth}
        \centering
        \vspace{0.15cm}
        \includegraphics[width=0.85\linewidth]{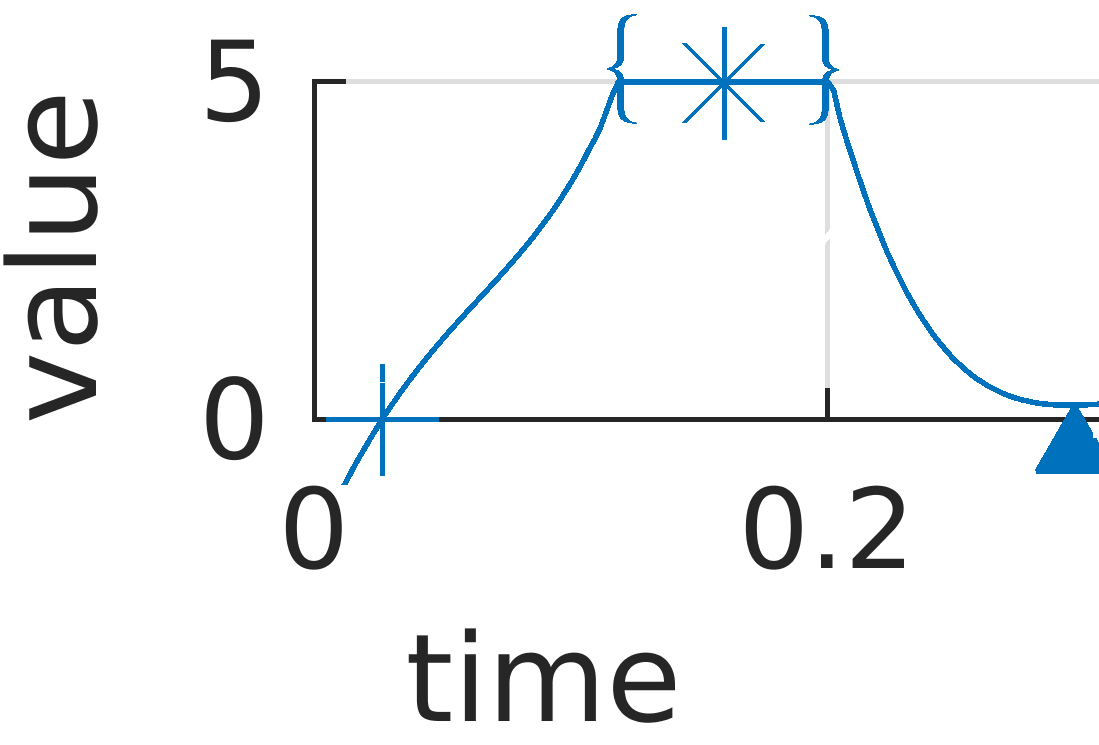}
        \caption{The two events \qq{start of constrained arc} ($\lbrace$) and \qq{end of constrained arc} ($\rbrace$) and are merged into a single event \qq{constrained arc} ($\varhexstar$) in the center of the constraint.}
        \label{fig:example_boxconstrAsMax}
    \end{minipage}
    \hfill
    \vline
    \hfill
    \begin{minipage}{0.22\textwidth}
        \centering
        \vspace{0.15cm}
        \includegraphics[width=0.95\linewidth]{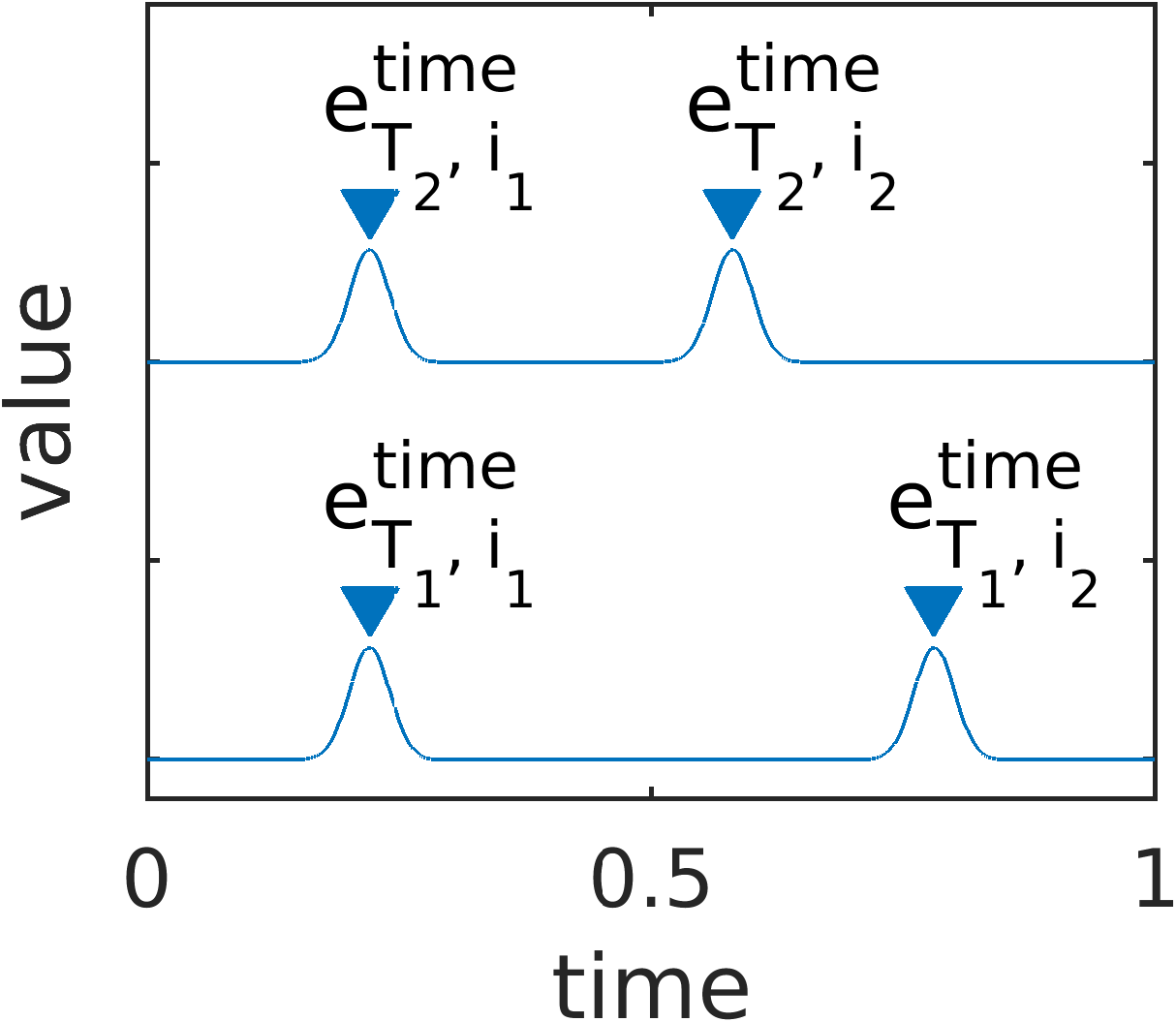}
        \caption{Example of two simple trajectories, both have two features each, in different temporal distance.
        }
        \label{fig:exampletimegaps}
    \end{minipage}
\end{figure}

To find the active box constraint feature in linear splines $c$, we again consider only knots at times $t_i$ and check if the condition
\begin{equation}\label{eq:limit_condition}
u_c - c(t_i) < \varepsilon \text{ or } c(t_i) - l_c < \varepsilon
\end{equation} is fulfilled, with a small threshold $\varepsilon>0$. 
For cubic splines, we also check the condition \eqref{eq:limit_condition} at knots and extrema and proceed as for linear splines. 
Depending on whether the upper or lower bound is active, we use different attributes ($L_u$, $L_l$) in $e_i^\text{cat}$. Further, one could distinguish between a single touch point and the two points start and end of a constrained arc. In this work, we use a single feature for both cases and set the time $e_i^\text{time}$ of an arc to the center point between the start and end of the active constraint (see Figure \ref{fig:example_boxconstrAsMax} for an example). 
To get salience values $e_i^\text{val}$, we treat active box constraints as extrema (which they actually are) and compute the prominence value at these points.

\subsubsection{Roots}
\begin{figure}
    \vspace{0.15cm}
    \centering
    \subfloat[Trajectory graph with features indicated by symbols: \rotatebox{90}{$\blacktriangleright$} for minima, \rotatebox{90}{$\blacktriangleleft$} for maxima, $+$ for roots and $\varhexstar$ for active constraints.]{
        \includegraphics[width=0.85\linewidth]{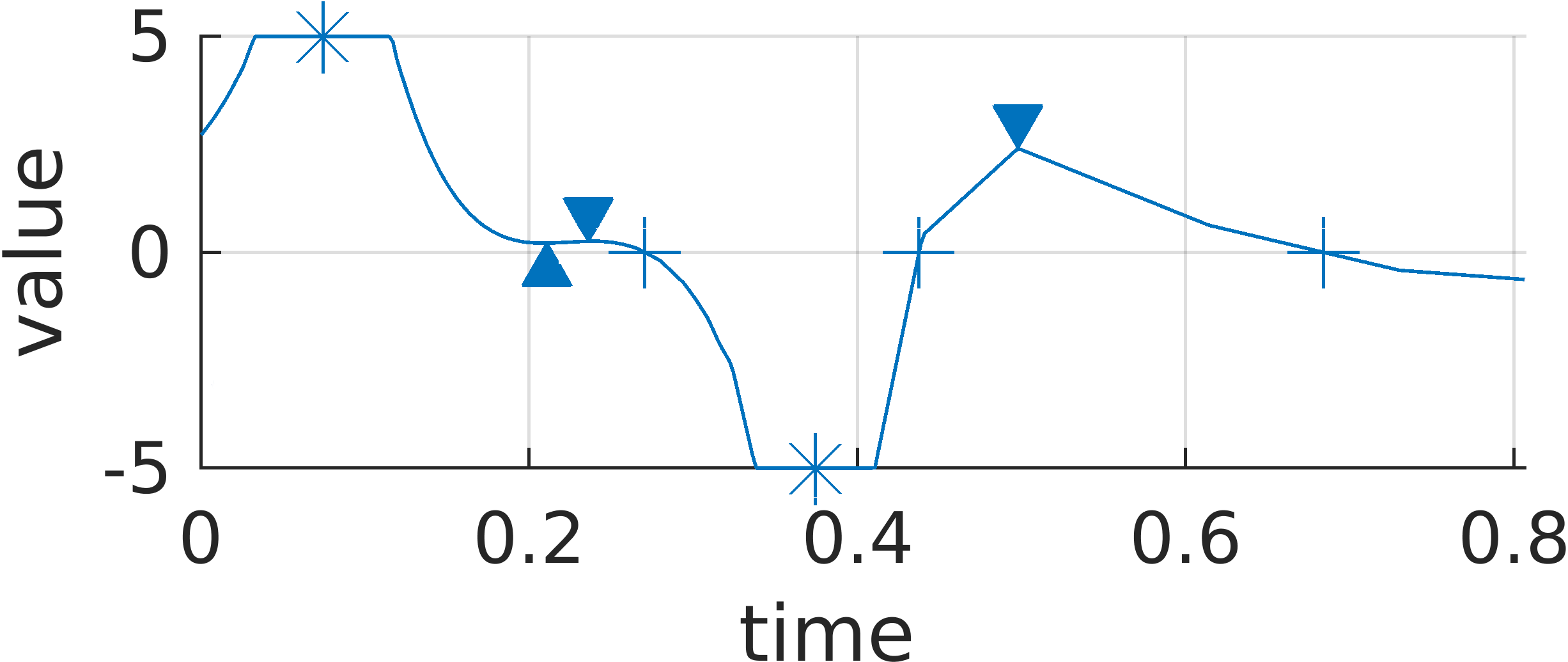}
    }\\
    \subfloat[Feature sequence for the above trajectory graph. Feature labels: $\wedge$: maximum, $\vee$: minimum, $0$: root, $L_u$/$L_l$: upper/lower box constraint active.]{
        \footnotesize
        \begin{tabularx}{0.975\linewidth}{@{}Xcccccccc@{}}
            \toprule
            $i$ & 1 & 2 & 3 & 4 & 5 & 6 & 7 & 8\\\midrule
            $ e_i^\text{cat} $ & $L_u$ & $\vee$ & $\wedge$ & $0$ & $L_l$ & $0$ & $\wedge$ & $0$\\
            $ e_i^\text{time} $ & 0.07 & 0.21 & 0.24 & 0.27 & 0.37 & 0.44 & 0.50 & 0.68\\
            $ e_i^\text{val} $ & 0.23 & 0.01 & 0.01 & 0.97 & 0.74 & 1.00 & 0.19 & 0.93\\
            \bottomrule
        \end{tabularx}
    }
    \caption{Example showing the representation of a trajectory graph with a feature sequence.}\label{fig:fseqexample}
\end{figure}

Considering roots can be helpful for motions where symmetry is relevant.
The roots ($e_i^\text{cat}$ indicated by $0$) are identified by finding the roots of all (linear or cubic) polynomials that are part of the trajectory. The value of each root at $e^\text{time}_i$ is computed using the slope $m =\frac{\text{d}}{\text{d}t}s(e^\text{time}_i)$ as   
\begin{equation}\label{eq:salience_roots}
e^\text{val}_i = 2\pi^{-1}\left|\arctan\left(m\right)\right|.
\end{equation}
Eq. \ref{eq:salience_roots} assigns salience values close to one to roots that intersect the x-axis with large slope and values close to zero for roots with flat angles.
For all roots at knots of linear splines, we use the subderivative $m = \frac{1}{2}(m^++m^-)$, where $m^-$ and $m^+$ denote the left and right limit, as they are not differentiable at these points.
An example of a feature sequence for a one-dimensional trajectory, including time and salience values, is given in Figure \ref{fig:fseqexample}.

\subsection{Step 2: Distance Metric for Feature Sequences}\label{sec:distance_metric}
For two motion plans $T_1$ and $T_2$, we construct the feature sequences as described in Section \ref{sec:representation_construction}. For these sequences, a distance needs to be computed to compare $T_1$ and $T_2$. We do this using the measure based on the string kernel method described in \cite{elzinga_sequence_2003, elzinga_versatile_2012}, which allows incorporation of the features $e^\text{time}$ and $e^\text{val}$. This measure is a metric, which means in particular that the triangular inequality holds, which \qq{ensures coherence between computed dissimilarities} \cite{studer_what_2016}.

In the following, we briefly summarize Elzinga's distance measure \emph{SVRspell} \cite{elzinga_sequence_2003, elzinga_versatile_2012}.
Consider a finite alphabet $\Sigma$ and the set $\Sigma^*$ of all possible strings (character sequences) that can be built from $\Sigma$. Enumeration of the elements in $\Sigma^*$ gives an infinite-dimensional vector space $S$ where each entry corresponds to one string.
For two strings $s_1, s_2 \in \Sigma^*$ the two corresponding vectors $r_{s1}, r_{s2}\in S$ list the number of respective sub-sequences that can be found in the strings. A short example is given in Table \ref{example_elzinga_space}.

\begin{table}
    \centering
    \vspace{0.15cm}
    \begin{tabularx}{\linewidth}{@{}XrlX@{}}
        \toprule
        &$\Sigma$ = $\left\lbrace a, b\right\rbrace$, &  $\Sigma^* = \left\lbrace a, b, aa, ab, ba, bb, aaa, \dots\right\rbrace$&\\[0.2cm]
        &$s = abb$, & $r_{s}=\begin{pmatrix}1&2&0&2&0&1&0&\dots\end{pmatrix}$&\\
        \bottomrule
    \end{tabularx}
    \caption{Example of a string $s$ and its vector $r_{s}$ denoting how often all sub-sequences occur. In this example, the sub-sequences in $s$ are \emph{a}, \emph{b}, \emph{ab}, \emph{bb}, \emph{abb}.}\label{example_elzinga_space}
    \vspace{-0.15cm}
\end{table}

To get the distance between $s_1$ and $s_2$, Elzinga et al. compute the Euclidean distance between $r_{s1}$ and $r_{s2}$. This can be formulated using only inner products, which can be computed even if the vectors are infinite-dimensional using so-called \emph{kernels}:
\begin{align}\label{eq_elzinga_def}
d(s_1, s_2) &= \sqrt{\sum_{i\in\N}\left(r_{s1,i}-r_{s2,i}\right)^2} \\ &= \sqrt{r_{s1}^Tr_{s1} + r_{s2}^Tr_{s2} - 2r_{s1}^Tr_{s2}}.\notag
\end{align}

\noindent
In \cite{elzinga_versatile_2012}, several extensions of SVRspell are described, some of them are used in this work. 
In the following, we explain how we use this measure and its extensions to allow the application on our representation of motion plans from \ref{sec:representation_construction}.

The alphabet in \cite{elzinga_versatile_2012} corresponds to our set of trajectory feature classes, a string with our feature sequence, and a character with a feature $e_i^\text{cat}$.
We can compute the distance between motion plans by applying \eqref{eq_elzinga_def} to each dimension of our feature sequence representation of the state and control trajectories.
The resulting $n+m$ distance values $d_i$ are combined into a single distance measure by taking the root of the sum of squared distances:
\begin{equation}
d_\text{final} = \left(\sum\nolimits_{i=1}^{n+m}{d_i}\right)^{-\frac{1}{2}}
\end{equation}
The following extensions of \cite{elzinga_versatile_2012} are used to include $e^\text{time}$ and $e^\text{val}$ in the distance measure.

Elzinga et al. describe how to add a concept named \emph{soft-matching} to the string kernel, which allows quantification of similarity between distinct characters of the alphabet. As an example, consider the alphabet $\left\lbrace a,b,c\right\rbrace$ and the soft-matching given by the similarity $0.7$ between $a$, $b$ and $0.0$ between both $a$, $c$ and $b$, $c$. Then, the distance between words $ab$ and $aa$ is smaller than between the words $ac$ and $aa$.

\noindent 
We make use of soft-matching to take into account the similarity of extrema and box constraints. The value that gives the similarity between $\wedge$ and $L_u$ or $\vee$ and $L_l$ is between $0$ and $1$, care must be taken that the resulting soft-matching matrix is positive definite.

The \emph{Trail-algorithm} in \cite{elzinga_versatile_2012} allows the introduction of a penalty term to decrease the influence of sub-sequences with large gaps. Gaps are differences between indices where a character of a sub-sequence occurs in a string. For example, the sub-sequence \emph{abc} of the string \emph{abac} has gaps $1$ and $2$ between \emph{ab} and \emph{ac}, respectively.
This gap penalty can be an arbitrary data-dependent function, which allows us to incorporate the normalized trajectory times to penalize differences in the time spans between two features in their trajectories, as exemplified in Figure \ref{fig:exampletimegaps}.
To measure distances between feature sequences in our approach, we use the following gap weighting function:
\begin{equation}\label{eq:gap_penalty_fcn}
g(i_1, j_1, i_2, j_2) = 1-\left|\left(e_{T_1, i_2}^\text{time}-e_{T_1, i_1}^\text{time}\right) - \left(e_{T_2, j_2}^\text{time}-e_{T_2, j_1}^\text{time}\right)\right|.
\end{equation}
The inputs $i_1$ and $i_2$ are indices of two features in the trajectory $T_1$ and $j_1$, $j_2$ are indices for $T_2$.
In the kernel, each entry of $r_{s}$ of a feature sequence is weighted by a product of the gap weights computed by $g$ in the respective sub-sequence (slightly simplified). How to handle the case where a sub-sequence occurs more than once in a string is described more detailed in \cite{elzinga_versatile_2012}.

Finally, we need to add the information provided by the salience $e^\text{val}$ into the distance measure. In \cite{elzinga_versatile_2012}, Elzinga et al. propose an adaptation of their method to cover so-called \emph{run-length encodings} of strings where repeated characters are encoded as numbers (e.g., \emph{aaabccccb} as \emph{$\text{a}^3\text{b}^1\text{c}^4\text{b}^1$}). This can also be used for \qq{any quantifiable property of the characters}, which is the salience $e^\text{val}$ in our application. Accordingly, each sub-sequence of a string is weighted by the sum of such a run-length. However, products of the weights are more appropriate for our application. This ensures that the importance of sub-sequences containing features with very little salience is reduced as a whole. This extension requires a single line change in the existing algorithm: In the notion of \cite{elzinga_versatile_2012}, it suffices to modify line 5 of Elzinga's Grid-algorithm to assign \texttt{$m_{ij}^1 \leftarrow t_{xi}t_{yj}$}.

All in all, \emph{SVRspell} modified with three presented extensions (two of them already proposed in \cite{elzinga_versatile_2012}, one implemented for this work) provides a distance metric to compare feature sequences that represent motion plans. It allows incorporation of the additional information provided by temporal relation $e^\text{time}$ and salience $e^\text{val}$ of the features.

\subsection{Step 3: Application of Hierarchical Clustering}
The distance measure described in the previous section enables us to construct a distance matrix for a given set of motion plans, for which the distance between each pair in this set needs to be computed. This distance matrix is required by the agglomerative hierarchical clustering algorithm that constructs the set into subsets of similar motion plans. The merging of clusters is done using single-linkage. The number of resulting clusters is either pre-defined or results from a cutoff value. Both values must be tuned, which is facilitated by considering a dendrogram.

\section{Evaluation and Experiments}\label{sec_evaluation}
We compare the feature-based distance measure of motion plans based on SVRspell with DTW on the raw pairs of state and control trajectories. We selected DTW because it is, according to Su et al., \qq{one of the most widely used distance measures} \cite{su_survey_2020}.
Computations are performed in Matlab R2020a on a laptop with an Intel i7-6500U processor with two cores at 2.5 GHz. We use the DTW implementation of the Matlab Signal Processing Toolbox (compiled code) and the agglomerative hierarchical clustering from the Matlab Statistics and Machine Learning Toolbox. The SVRspell algorithm is implemented in C++ and used in Matlab via MEX file. The feature extraction proposed in this paper is written in Matlab code.
The sets of trajectories that are used in this evaluation are, if not stated otherwise, provided by the optimal control problem solver DIRCOL \cite{stryk_numerical_1993}.

\subsection{Clustering of Furuta Pendulum Motion Plans}
The underlying physical system of the motion plans in this subsection is the Furuta pendulum \cite{furuta_swing_1992}. It is an underactuated rotary pendulum with two rotational joints, where the first joint can be actuated and the second joint rotates freely.
An optimal control problem is formulated to get the controls for a swing-up motion that brings the passive second arm from a hanging start state into an upright position. The objective function favors solutions of minimum time and minimum energy. The problem formulation contains no explicit information about trajectory clusters or distance measures.
Computing solutions of this optimal control problem from different start states sampled closely around the hanging state using the numerical solver DIRCOL gives a set of motion plans. The differences in the motion plans originate from the differences in the start states; the results provided by DIRCOL can be clustered into visually different solutions.
By changing the parameterization of the dynamic model, we get three different test sets \emph{Furuta 1} to \emph{3} of 160 motion plans each that are clustered separately.
We manually cluster the 160 trajectories for each test set and use this as ground truth. Trajectories with visually similar graphs are grouped into one cluster. The ground truth clustering has been done before developing our clustering approach.

For this problem, the \emph{root} feature is used for the state trajectories as the roots of the second joint indicate the number of swing-up motions.
A qualitative evaluation is performed by comparing the dendrograms and clusters resulting from DTW and our method. 
Each motion plan in the test set is listed on the x-axis of the dendrogram, the distances between clusters are on the y-axis.
The trajectory plots show the distinct clusters of the motion plans, which contain state (consisting of two joint positions and two velocities) and control trajectories. 
The qualitative evaluation is supported by comparing the number of correctly clustered motion plans as quantitative evaluation. The separation into clusters depends on the cutoff value chosen. Lower cutoff values lead to fragmentation into more clusters. Misclassifications with high distance to the other elements make a lower cutoff value necessary to separate the clusters.
The results for both methods are given for the separation into the most favorable number of clusters to allow a fair comparison. This means finding a reasonable tradeoff between the number of clusters and the lowest number of misclassifications.

\begin{figure}
    \centering
    \subfloat[Result with Feature-based\label{fig:Perm0_dFB}]{%
        \includegraphics[width=0.23\textwidth]{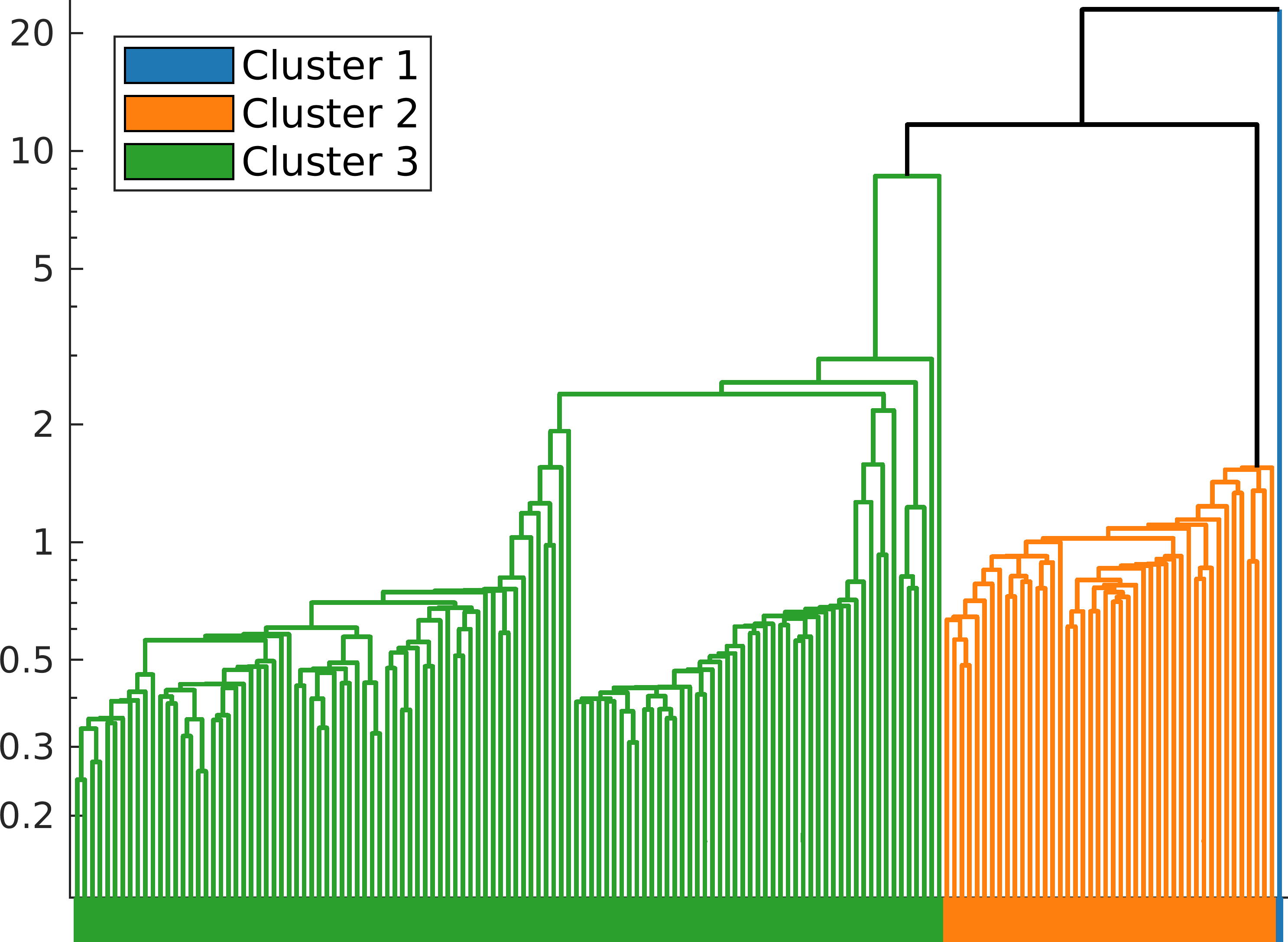}
    }
    \hfill
    \subfloat[Result with DTW\label{fig:Perm0_dDTW}]{%
        \includegraphics[width=0.23\textwidth]{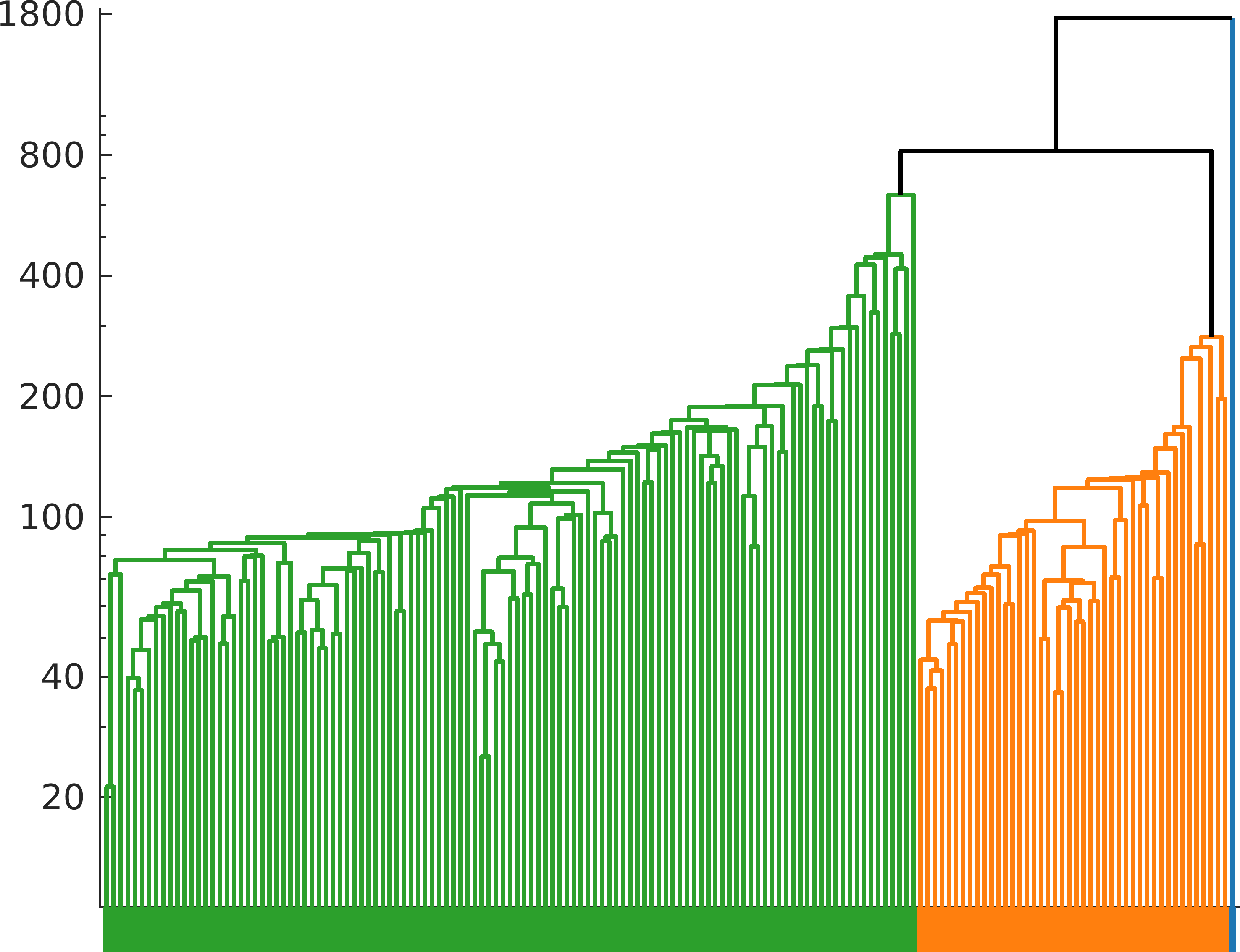}
    }
    \caption{The hierarchical structure of the clustering of test set \emph{Furuta 1} using different distance measures. The ground truth is given as colored bar on the x-axis.}
\end{figure}

The first test set \emph{Furuta 1} has three different clusters containing $44$, $115$ and $1$ trajectories. Both DTW and our method identify all clusters correctly (see Table \ref{quantitativeEval}). The dendrograms (Fig. \ref{fig:Perm0_dFB}, \ref{fig:Perm0_dDTW})
illustrate the subdivision into three clusters.
The second test set \emph{Furuta 2} has four different ground truth clusters with $90$, $38$, $1$ and $31$ trajectories. 
Our feature-based approach identifies the four clusters apart from two elements (Table \ref{quantitativeEval}).
DTW separates three single trajectories from the ground truth clusters and thus needs a lower cutoff value, leading to seven clusters.
Supporting the quantitative result, the dendrogram for the feature-based approach is structured more clearly: The four clusters are separated and the more fine-grained splitting happens only at a lower cutoff level. 

\begin{figure}
    \subfloat[\emph{Furuta 2}: Result Feature-based\label{fig:Perm1_dFB}]{%
        \includegraphics[width=0.23\textwidth]{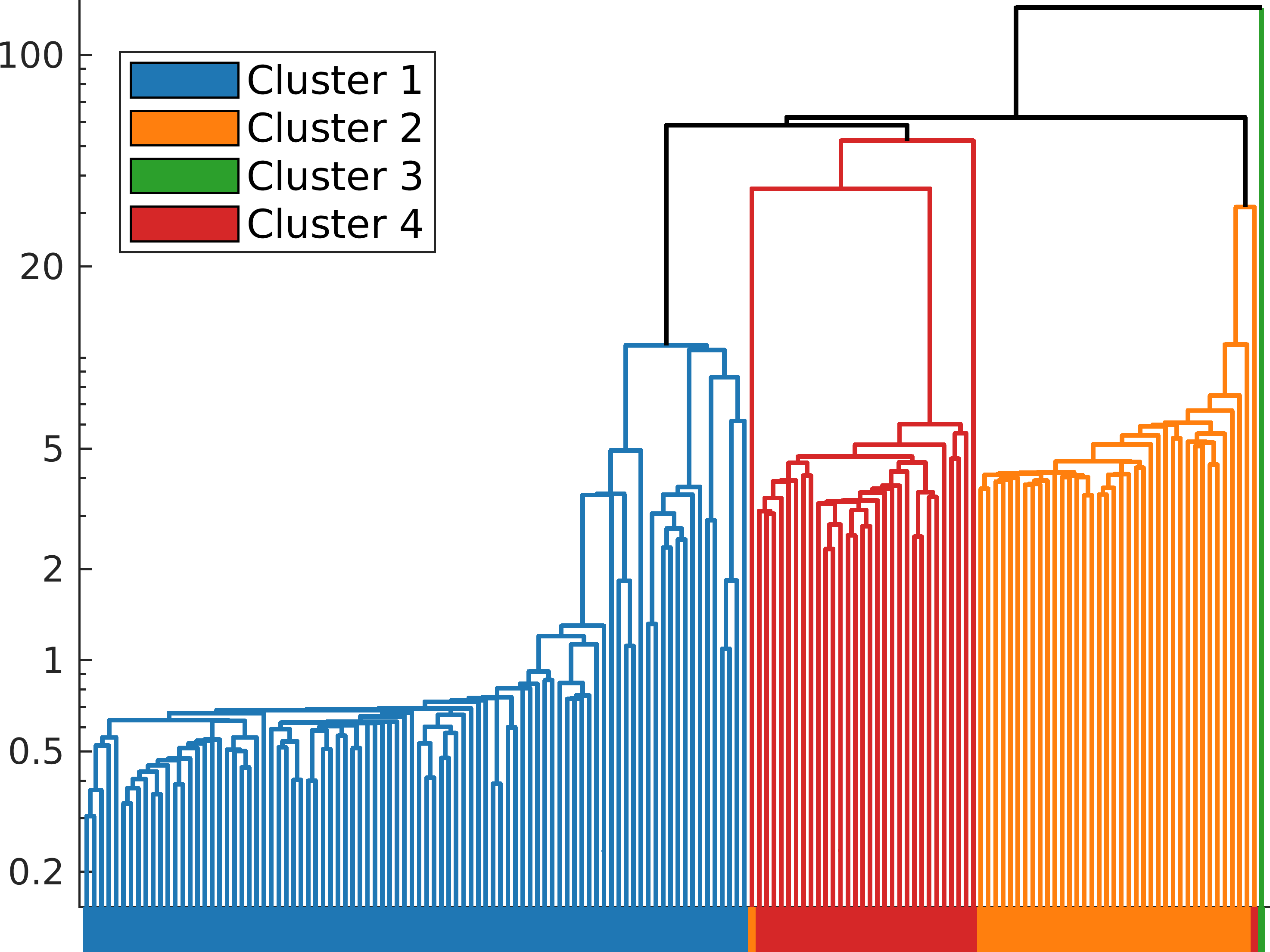}
    }
    \hfill
    \subfloat[\emph{Furuta 2}: Result with DTW\label{fig:Perm1_dDTW}]{%
        \includegraphics[width=0.23\textwidth]{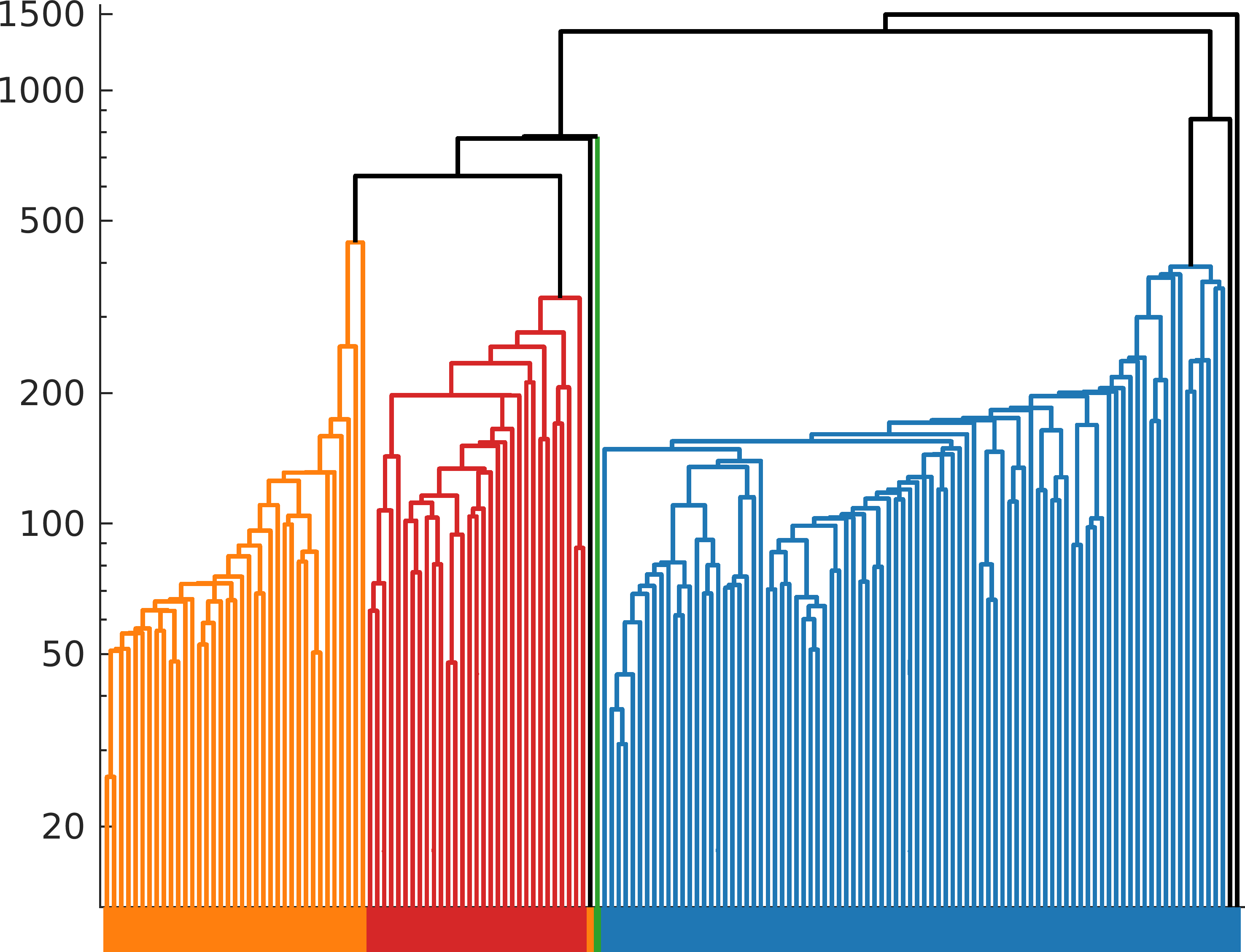}
    }
    \\
    \subfloat[\emph{Furuta 3}: Result Feature-based\label{fig:Perm2_dFB}]{%
        \includegraphics[width=0.23\textwidth]{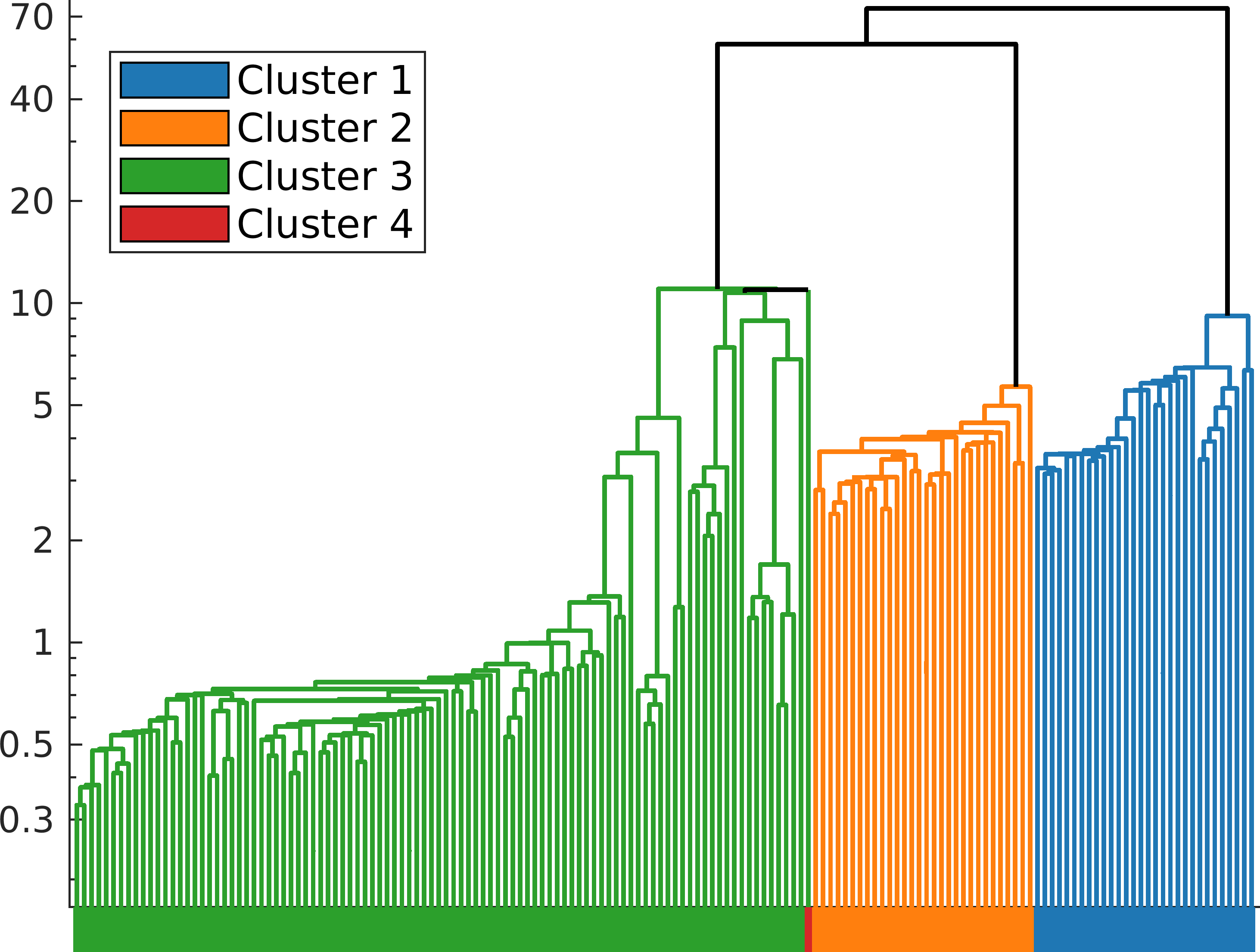}
    }
    \hfill
    \subfloat[\emph{Furuta 3}: Result with DTW\label{fig:Perm2_dDTW}]{%
        \includegraphics[width=0.23\textwidth]{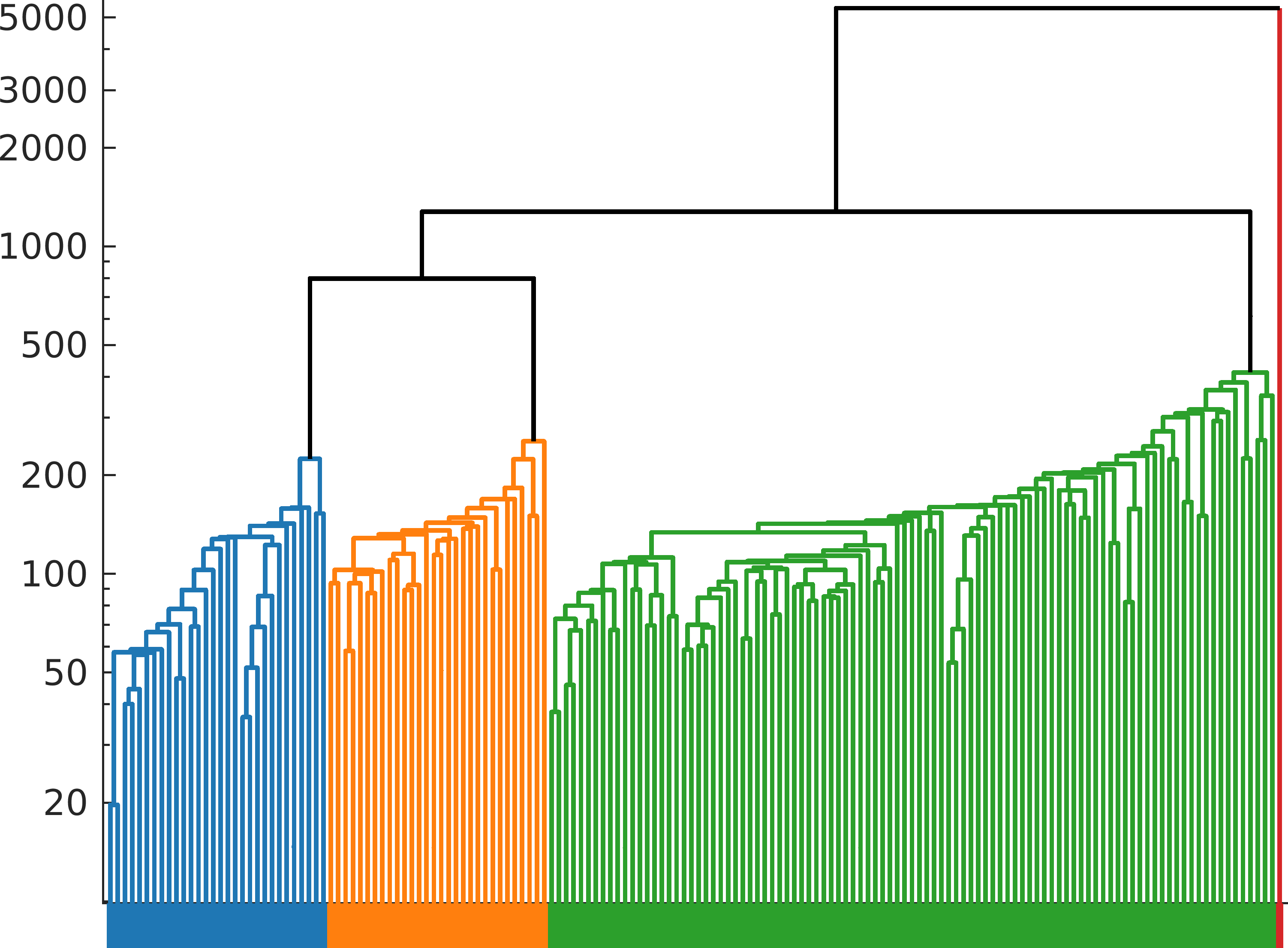}
    }
    \caption{
        The hierarchical structure of the clustering of test sets \emph{Furuta 2} and \emph{Furuta 3} using different distance measures.
    }\label{fig:testset_2}
\end{figure}

The third test set \emph{Furuta 3} consists of four trajectory clusters with sizes $30$, $30$, $99$ and $1$. On this test set, the clustering computed with DTW equals the manual clustering, the feature-based distance measure assigns the single trajectory cluster to the largest cluster. 

\begin{table*}
    \centering
    \vspace{0.14cm}
    \tabcolsep=5.7pt
    \begin{tabularx}{\linewidth}{@{}c|cccccc|ccccccccc|ccccccc@{}}
        \toprule
        \multicolumn{1}{l|}{} & \multicolumn{6}{c|}{\emph{Furuta 1}}                                                                         & \multicolumn{9}{c|}{\emph{Furuta 2}}                                                                                                  & \multicolumn{7}{c}{\emph{Furuta 3}}                                                                                                    \\
        \multicolumn{1}{l|}{GT} & \multicolumn{3}{c|}{Feature-based (3)}                       & \multicolumn{3}{c|}{DTW (3 clusters)}   & \multicolumn{4}{c|}{Feature-based (4)}                                    & \multicolumn{5}{c|}{DTW (7 clusters)}               & \multicolumn{3}{c|}{Feature-based (3)}                                    & \multicolumn{4}{c}{DTW (4 clusters)}                 \\[0.1cm]
        & $C_1$     & $C_2$     & \multicolumn{1}{c|}{$C_3$}& $C_1$      & $C_2$       & $C_3$ & $C_1$      & $C_2$       & $C_3$        & \multicolumn{1}{c|}{$C_4$}             & $C_1$       & $C_2$       & $C_3$      & $C_4$      & $\sdots$   & $C_1$       & $C_2$       & \multicolumn{1}{c|}{$C_3$}       & $C_1$       & $C_2$       & $C_3$       & $C_4$      \\ \midrule
        $C_1$                   & \textbf{1} &             & \multicolumn{1}{c|}{}             & \textbf{1} &             &              & \textbf{90} &             &            & \multicolumn{1}{c|}{}            & \textbf{88} &             &            &            & 2          & \textbf{30} &             & \multicolumn{1}{c|}{}           & \textbf{30} &             &             &            \\
        $C_2$                   &            & \textbf{44} & \multicolumn{1}{c|}{}             &            & \textbf{44} &              &             & \textbf{37} &            & \multicolumn{1}{c|}{1}           &             & \textbf{37} &            &            & 1          &             & \textbf{30} & \multicolumn{1}{c|}{}           &             & \textbf{30} &             &            \\
        $C_3$                   &            &             & \multicolumn{1}{c|}{\textbf{115}} &            &             & \textbf{115} &             &             & \textbf{1} & \multicolumn{1}{c|}{}            &             &             & \textbf{1} &            &            &             &             & \multicolumn{1}{c|}{\textbf{99}}&             &             & \textbf{99} &            \\
        $C_4$                   & --         & --          & \multicolumn{1}{c|}{--}           & --         & --          & --           &             & 1           &            & \multicolumn{1}{c|}{\textbf{30}} &             &             &            & \textbf{31}&            &             &             & \multicolumn{1}{c|}{1}          &             &             &             & \textbf{1} \\ \bottomrule
    \end{tabularx}
    \caption{The clusterings of the Furuta test sets as confusion matrices using the feature-based approach and DTW. The rows give the ground truth clusters, the columns the respective cluster assignments resulting from the two distance measures.  The number of clusters is given as number behind the method name. Additional clusters are summarized in \qq{$\sdots$}.
    }\label{quantitativeEval}
\end{table*}

\subsection{Clustering of Manutec r3 Arm Motion Plans}
In this subsection, we consider a more complex robotic system. The Manutec r3 robot arm model has three actuated joints, such that a motion plan consists of a six-dimensional state and a three-dimensional control trajectory. The motion plans that describe optimal point-to-point movements from different start states to the same goal state are computed again using the solver DIRCOL.
The test set based on the Manutec arm consists of 30 motion plans that are clustered manually into 9 different movements. There are three larger clusters of 10, 6 and 5 motions, respectively.
The other clusters contain one or two trajectories and are neglected in this evaluation.

A property of motion plans resulting from this problem is a low-amplitude zig-zag behavior in one dimension of the control trajectory. This validates our use of a small prominence filter (a salience threshold of 0.02 is sufficient) for the extrema to remove the many maxima and minima. 
The prominence filter is thus important to reduce the length of the feature sequence. Moreover, it can compensate for some noise in the input data. The feature class \emph{root} provides no useful information for trajectories of the Manutec test set and is thus not used for the state or control trajectories.

\begin{figure}
    \vspace*{-0.3cm}
    \centering
    \subfloat[Result with Feature-based\label{fig:Manu_dFB}]{%
        \includegraphics[width=0.23\textwidth]{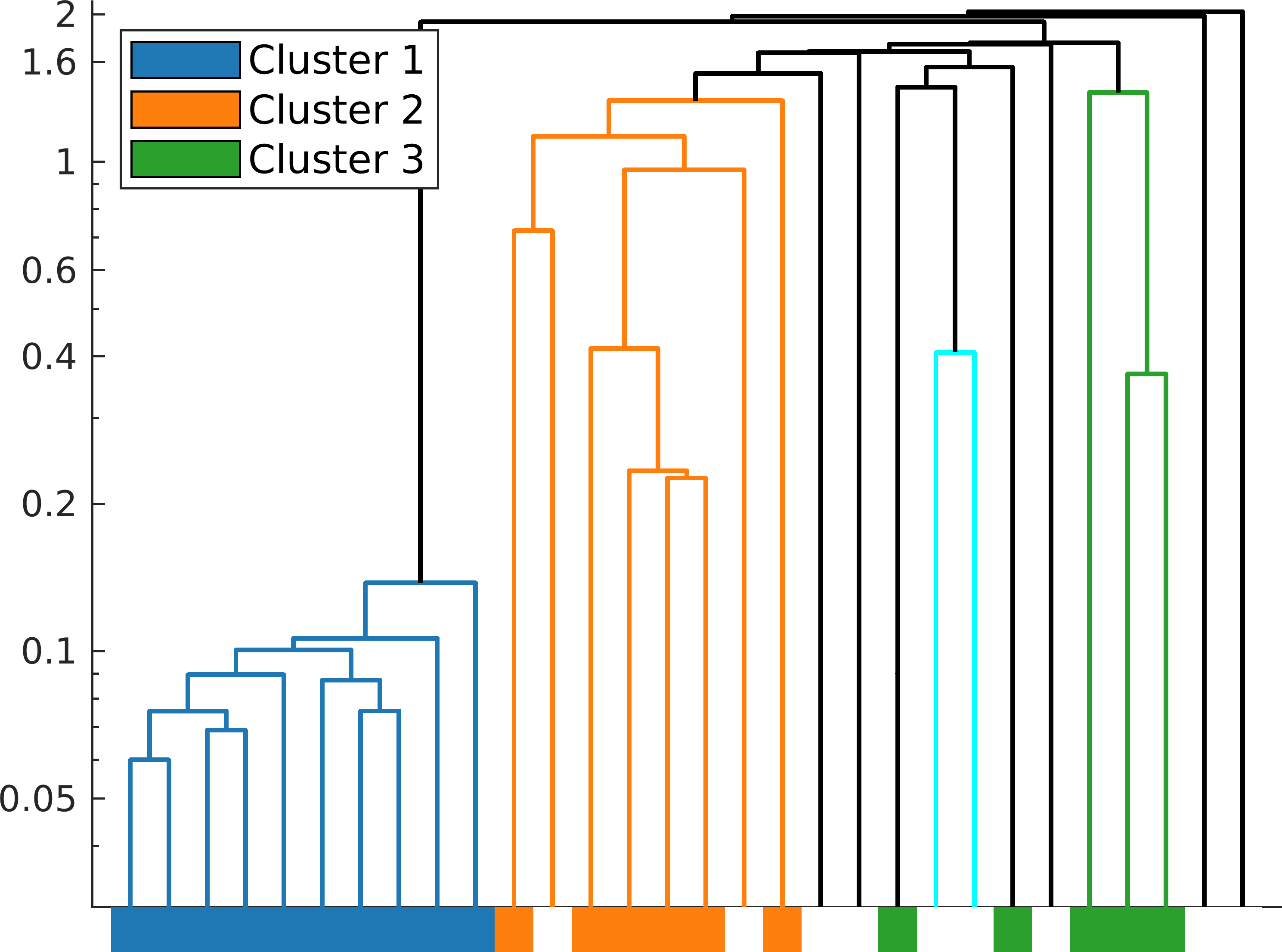}
    }
    \hfill
    \subfloat[Result with DTW\label{fig:Manu_dDTW}]{%
        \includegraphics[width=0.23\textwidth]{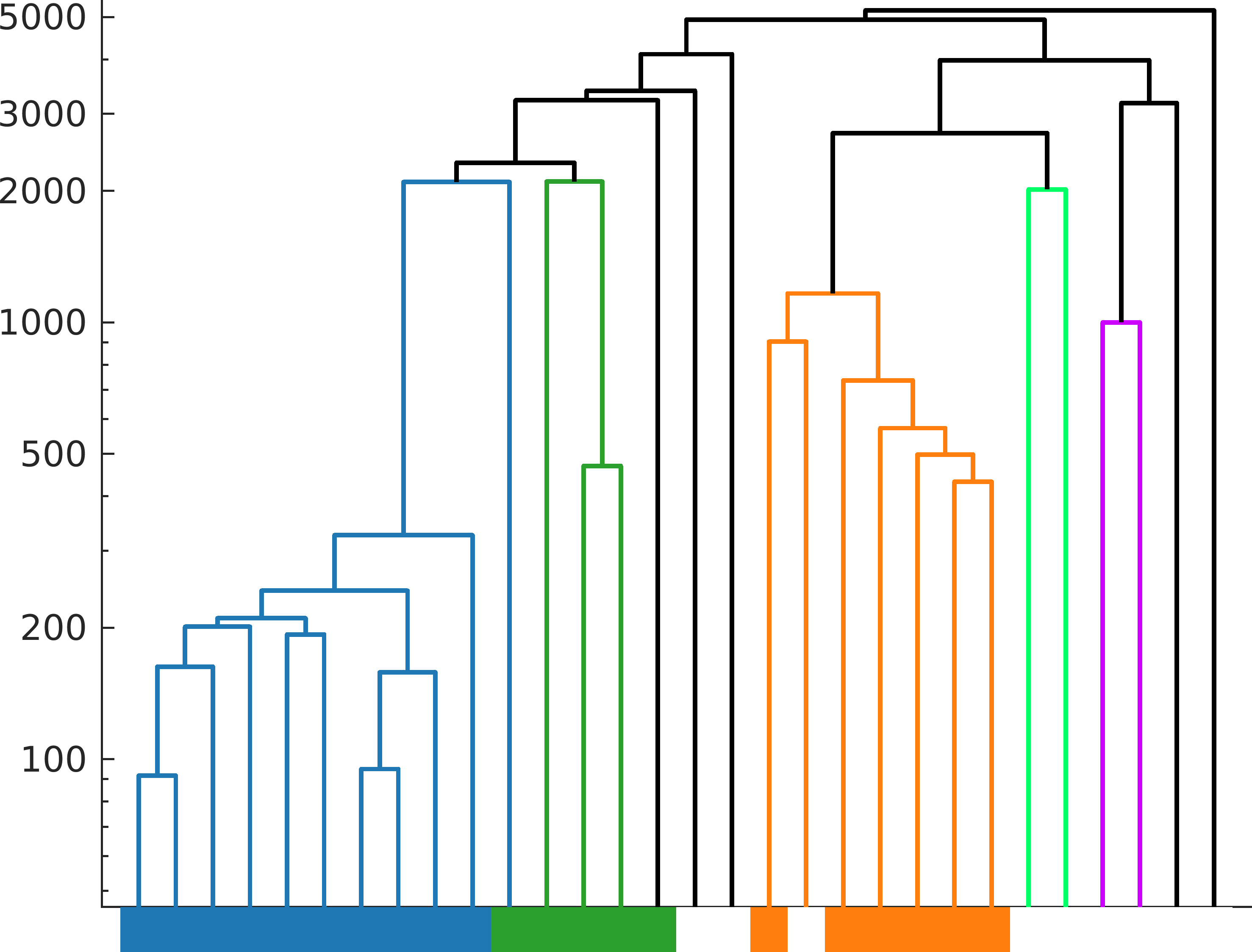}
    }
    \caption{The hierarchical structure of the clustering of test set \emph{Manutec} using different distance measures.}\label{fig:testsetManu}
\end{figure}

Table \ref{tab:resManutec} shows that, considering the number of correctly clustered motion plans in the three largest clusters, our feature-based approach has a slight advantage over DTW: The large cluster 1 is identified correctly, but cluster 3 is incomplete and the algorithm has problems separating the second cluster from other motion plans.
In contrast, DTW is unable to successfully identify any cluster: The second cluster includes one additional motion plan, the first cluster contains four from cluster 3.
Comparing the dendrograms (Fig. \ref{fig:testsetManu}), the diagram for DTW seems to be more clearly structured. The first cluster is presented contiguously, but a very low cutoff value is needed to separate it from other elements. In the dendrogram for our approach, this cluster is also clearly recognizable: It is separated from other motion plans, even for a very high cutoff value. Cluster 1 stands out visually from the other motion plans, it is clearly identified by the feature-based approach, whereas DTW has major problems with this cluster, as can be seen in the dendrogram.

All in all, both methods have some difficulties with the selected data set, which is reflected in the dendrograms. In several cases, the motion plans have similar progressions in single trajectories, which makes clustering challenging even for human experts. Interestingly, the two approaches identify different clusters well. However, considering the overall number of correctly classified motion plans, our feature-based approach has a slight edge on DTW.

\begin{table}
    \centering
    \begin{tabular}{@{}c|cccc|cccc@{}}
        \toprule
        GT   & \multicolumn{4}{c|}{Feature-based (11)}      & \multicolumn{4}{c}{DTW (10)}                 \\[0.1cm]
        & $C_1$       & $C_2$      & $C_3$      & $\sdots$& $C_1$    & $C_2$      & $C_3$      &$\boldmath\sdots$\\ \midrule
        $C_1$& \textbf{10} &            &            &      & \textbf{10} &            &            &      \\
        $C_2$&             & \textbf{6} &            &      &             & \textbf{6} &            &      \\
        $C_3$&             &            & \textbf{3} & 2    & 4           &            & \textbf{1} &      \\
        $\boldmath\sdots$&          & 3          &            & $\sdots$  &        & 1          &            & $\sdots$  \\ \bottomrule
    \end{tabular}
    \caption{The results of the clustering as confusion matrix for the Manutec robot arm test set comparing feature-based and DTW distance measures.
        Only the three largest clusters are considered, all others are summarized in \qq{$\sdots$}.} 
    \label{tab:resManutec}
\end{table}

\subsection{Clustering of a Real-world Human Motion Dataset}
To evaluate the performance of the presented method on real-world data, we cluster trajectories from the human locomotion dataset created by Reznick et al. \cite{reznick_lower_2021}. They recorded data from ten participants walking, running, stair climbing and sitting down/standing up at various speeds and inclinations. The recorded data is post-processed (filtered and gaps in data filled). We use the normalized dataset where the strides in the recorded trajectories are separated and the time is normalized to 1 and interpolated to 150 data points (for details, see \cite{reznick_lower_2021}).
In our evaluation, we use the provided joint angle trajectories for the ankle, knee, hip, pelvis and foot progression, and from each of them only the x-component which is the most expressive and thus provides the best results. Consequently, the state trajectories to be clustered have five dimensions. There is no control signal information, the joint limits are not known and roots have no meaning, such that we use only the state trajectory with maxima and minima as features. Extrema with prominence lower than 0.2 are removed from the feature sequence.

\begin{figure}
    \centering
    \subfloat[\emph{Human Motion 1}: Result with Feature-based approach\label{fig:Texas0_dFB}]{%
        \includegraphics[width=0.23\textwidth]{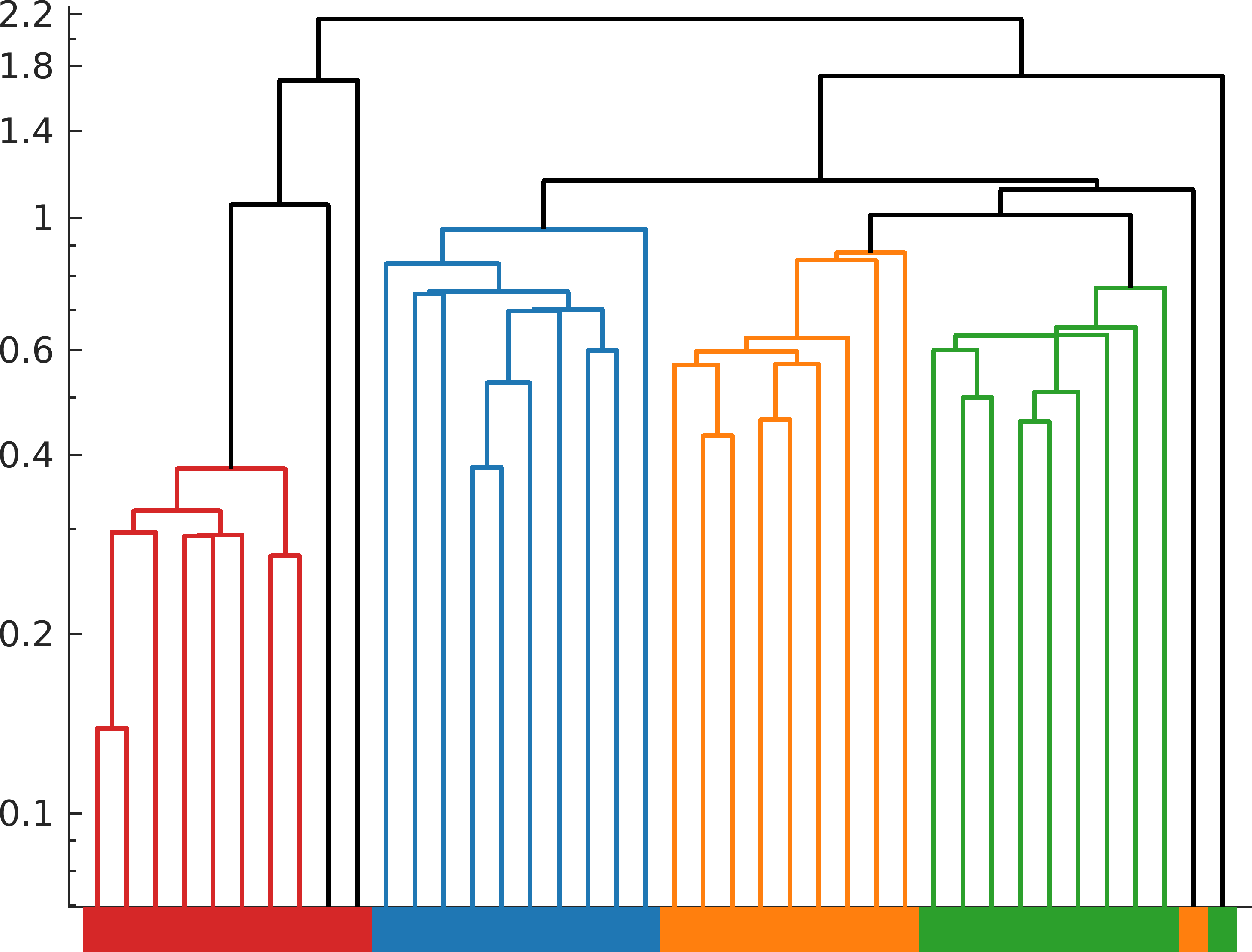}
    }
    \hfill
    \subfloat[\emph{Human Motion 1}: Result DTW\label{fig:Texas0_dDTW}]{%
        \includegraphics[width=0.23\textwidth]{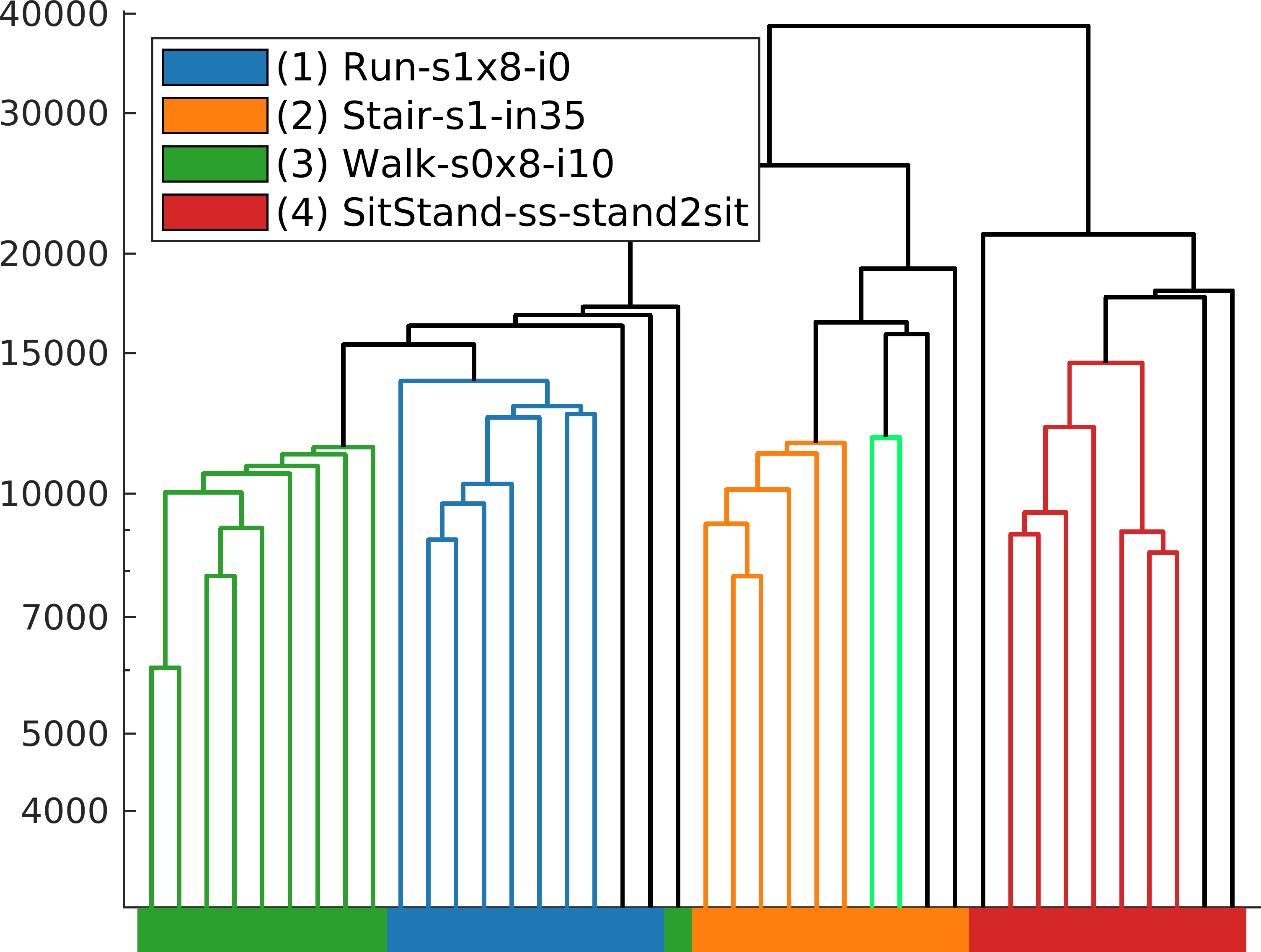}
    }
    \\
    \subfloat[\emph{Human Motion 2}: Result with Feature-based approach\label{fig:Texas1_dFB}]{%
        \includegraphics[width=0.23\textwidth]{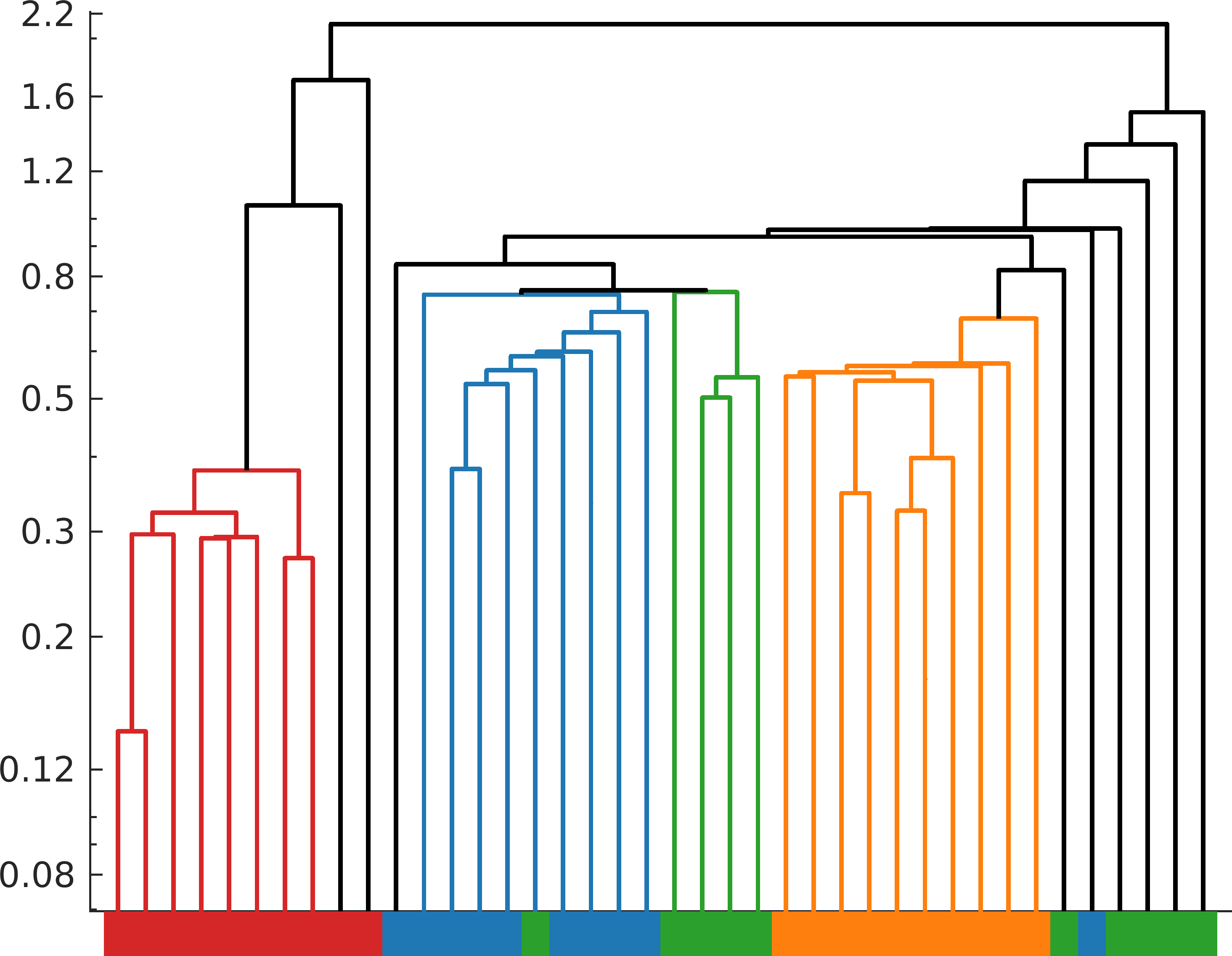}
    }
    \hfill
    \subfloat[\emph{Human Motion 2}: Result DTW\label{fig:Texas1_dDTW}]{%
        \includegraphics[width=0.23\textwidth]{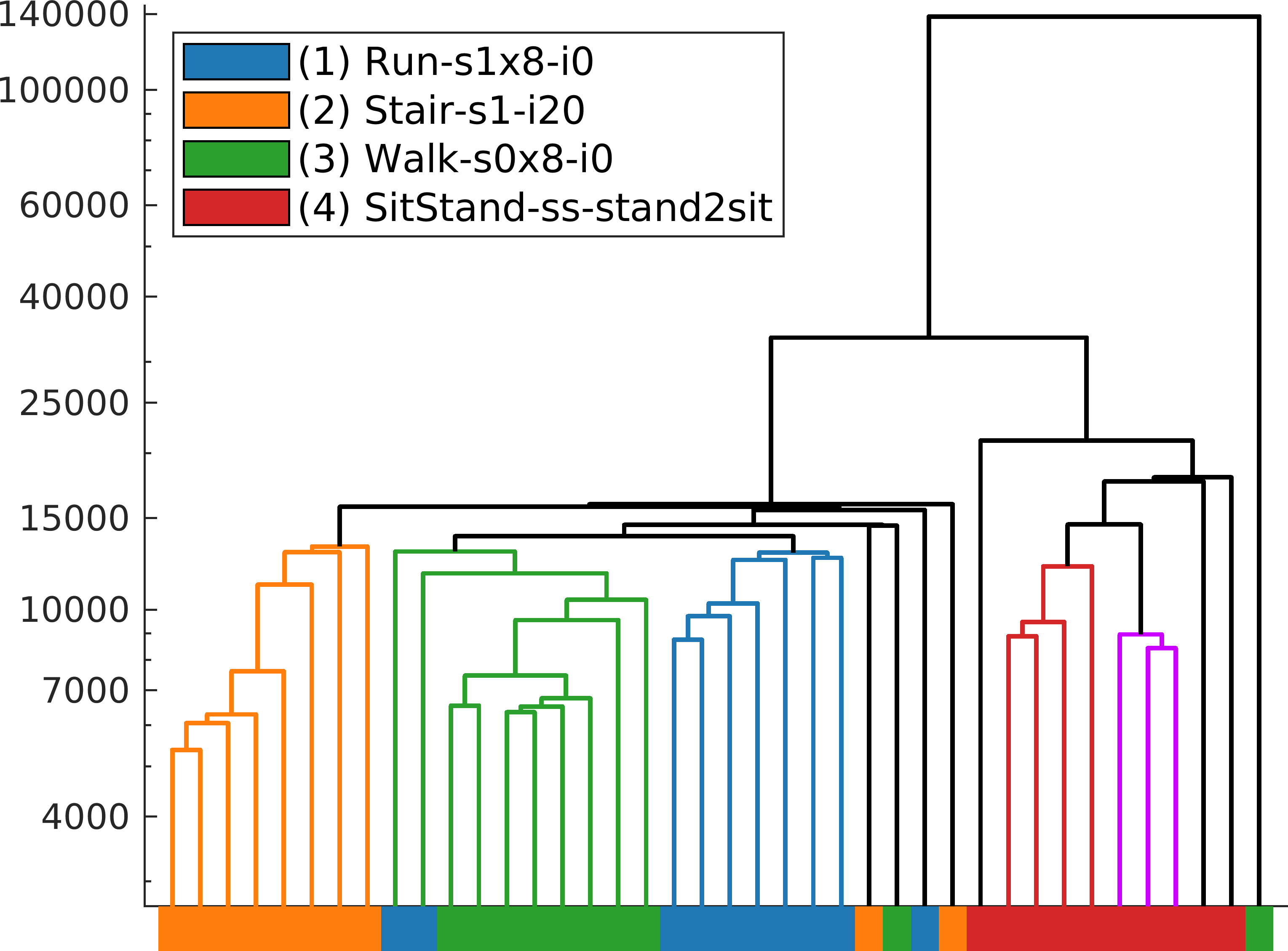}
    }
    \caption{The hierarchical structure of the clustering of test sets \emph{Human Motion 1} and \emph{Human Motion 2} using different distance measures. Colored bars represent ground truth.
    }\label{fig:Texas0}
\end{figure}

A cluster consists of the first stride of all ten participants in a single motion task.
The distinctness of the different motions in the data set varies, such that the difficulty of the clustering task varies with the selection of motions. For example, differentiating between joint trajectories for walking with different inclinations is also difficult for humans; both clustering methods fail to cluster these trajectory bundles. In the test set \emph{Human Motion 1}, we cluster into the following motion types which are better to differentiate: (1) running at \SI[per-mode=symbol]{1.8}{\meter\per\second}, (2) descending a \SI{35}{\degree} stair, (3) walking \SI{10}{\degree} downhill at \SI[per-mode=symbol]{0.8}{\meter\per\second} and (4) sitting down on a chair. The results of the clustering using DTW and the feature-based approach are presented in Fig. \ref{fig:Texas0_dFB}--\ref{fig:Texas0_dDTW}. By and large, all four clusters can be identified in the dendrograms. The feature-based approach has a problem to distinguish between the clusters (2) and (3), DTW to differentiate between (1) and (3).
Both approaches report high distance values for some trajectories of a motion type and thus separate them from the others, resulting in more than four clusters. As before, we report in Table \ref{tab:resHumanMotion} (upper) the separation into the most favorable number of clusters for the feature-based approach and for DTW two alternative separations into 13 and 4 clusters.
It can be seen that the four ground truth motions are identified with much higher cutoff values using the feature-based approach, and also the number of misclassifications is substantially lower.

\begin{table}
    \centering
    \vspace{0.14cm}
    \tabcolsep=0.11cm
    \begin{tabularx}{\linewidth}{@{}c?cccccccccccccc@{}}
        \toprule
        \multicolumn{1}{l}{}   & \multicolumn{14}{c}{Human Motion 1}                                                                                                                                                                                                                      \\
        \multirow{2}{*}{GT} & \multicolumn{5}{c?}{Feature-based (8)}                                        & \multicolumn{5}{c|}{DTW (13 clusters)}                                       & \multicolumn{4}{c}{DTW (4 clusters)}                                                      \\
        & $C_1$       & $C_2$      & $C_3$      & $C_4$      & \multicolumn{1}{c?}{...} & $C_1$      & $C_2$      & $C_3$      & $C_4$      & \multicolumn{1}{c|}{...} & \multicolumn{1}{l}{$C_1$} & \multicolumn{1}{l}{$C_2$} & \multicolumn{1}{l}{$C_3$} & $C_4$ \\ \midrule
        $C_1$                   & \textbf{10} &            &            &            & \multicolumn{1}{c?}{}    & \textbf{8} &            &            &            & \multicolumn{1}{c|}{2}   & \textbf{0}                &                           & 10                        &       \\
        $C_2$                   &             & \textbf{9} &            &            & \multicolumn{1}{c?}{1}   &            & \textbf{6} &            &            & \multicolumn{1}{c|}{4}   &                           & \textbf{10}               &                           &       \\
        $C_3$                   &             &            & \textbf{9} &            & \multicolumn{1}{c?}{1}   &            &            & \textbf{9} &            & \multicolumn{1}{c|}{1}   &                           &                           & \textbf{10}               &       \\
        $C_4$                   &             &            &            & \textbf{8} & \multicolumn{1}{c?}{2}   &            &            &            & \textbf{7} & \multicolumn{1}{c|}{3}   & 1                         &                           &                           & 9     \\ \bottomrule
    \end{tabularx}
    \newline\vspace*{0.1cm}\newline
    \tabcolsep=0.037cm
    \begin{tabularx}{\linewidth}{@{}c?cccccccccccccccccccc@{}}
        \toprule
        & \multicolumn{20}{c}{Human Motion 2}                                                                                                                                                                                                                                                      \\
        \multirow{2}{*}{GT} & \multicolumn{5}{c|}{Feature-bd. (13)}                                    & \multicolumn{5}{c?}{Feature-bd. (10)}                                    & \multicolumn{5}{c|}{DTW (13)}                                       & \multicolumn{5}{c}{DTW (8)}                \\
        & $C_1$      & $C_2$       & $C_3$      & $C_4$ & \multicolumn{1}{c|}{...} & $C_1$      & $C_2$       & $C_3$      & $C_4$ & \multicolumn{1}{c?}{...} & $C_1$      & $C_2$      & $C_3$      & $C_4$      & \multicolumn{1}{c|}{...} & $C_1$       & $C_2$      & $C_3$      & $C_4$ & ... \\ \midrule
        $C_1$ & \textbf{8} &             &            &       & \multicolumn{1}{c|}{2}   & \textbf{9} &             &            &       & \multicolumn{1}{c?}{1}   & \textbf{7} &            & 2          &            & \multicolumn{1}{c|}{1}   & \textbf{10} &            &            &       &     \\
        $C_2$ &            & \textbf{10} &            &       & \multicolumn{1}{c|}{}    &            & \textbf{10} &            &       & \multicolumn{1}{c?}{}    &            & \textbf{8} &            &            & \multicolumn{1}{c|}{2}   & 1           & \textbf{8} &            &       & 1   \\
        $C_3$ & 1          &             & \textbf{4} &       & \multicolumn{1}{c|}{5}   & 5          & 1           & \textbf{1} &       & \multicolumn{1}{c?}{3}   &            &            & \textbf{8} &            & \multicolumn{1}{c|}{2}   & 9           &            & \textbf{1} &       &     \\
        $C_4$ &            &             &            &\textbf{8}& \multicolumn{1}{c|}{2}&            &             &            &\textbf{8}& \multicolumn{1}{c?}{2}&            &            &            & \textbf{4} & \multicolumn{1}{c|}{6}   &             &            &            &\textbf{7}& 3\\ \bottomrule
    \end{tabularx}
    \caption{Clustering results as confusion matrices for the Human motion dataset \cite{reznick_lower_2021}. Motion types (ground truth) given in rows, assigned clusters in columns. 
        Two separations into clusters are given if it is unclear which one is best.
        %
    }
    \label{tab:resHumanMotion}
    \vspace*{-0.08cm}
\end{table}

More difficult to differentiate are the motion types of \emph{Human Motion 2}: (1) running at \SI[per-mode=symbol]{1.8}{\meter\per\second}, (2) ascending a \SI{20}{\degree} stair, (3) walking at \SI[per-mode=symbol]{0.8}{\meter\per\second} and (4) sitting down on a chair. The results are presented in Fig. \ref{fig:Texas1_dFB}--\ref{fig:Texas1_dDTW}. Here, the ground truth clusters are more difficult to distinguish in both dendrograms. The feature-based approach needs a separation into 13 clusters to distinguish four larger clusters, DTW also needs 13 clusters to distinguish between cluster (1) and (3). The results for different numbers of clusters are given in Table \ref{tab:resHumanMotion} (lower). All in all, the performance of both approaches is comparable on this more difficult test set; both reveal difficulties in differentiating between cluster (1) and cluster (3) in particular.

\subsection{Evaluation of Efforts and Runtime}\label{sec_runningtimes}
The runtime analysis of the feature-based distance computation can be divided into the feature extraction step and the \emph{SVRspell}-based distance computation.
The feature extraction is a preprocessing step that must be done only once for all trajectories, independent of the number of distance measurements required for the clustering. Its runtime depends on the number of feature classes and the algorithms employed to identify the features of a trajectory. The complexity of the distance computation (based on the SVRspell algorithm) is cubic \cite{elzinga_versatile_2012} in the \emph{number of features}. The number of features is typically much smaller than the number of time steps, on which the complexity of DTW depends ($\mathcal{O}(NM)$ with $N$, $M$ the trajectories' number of time steps).

\begin{figure}
    \centering
    \vspace{0.15cm}
    \includegraphics[width=0.98\linewidth]{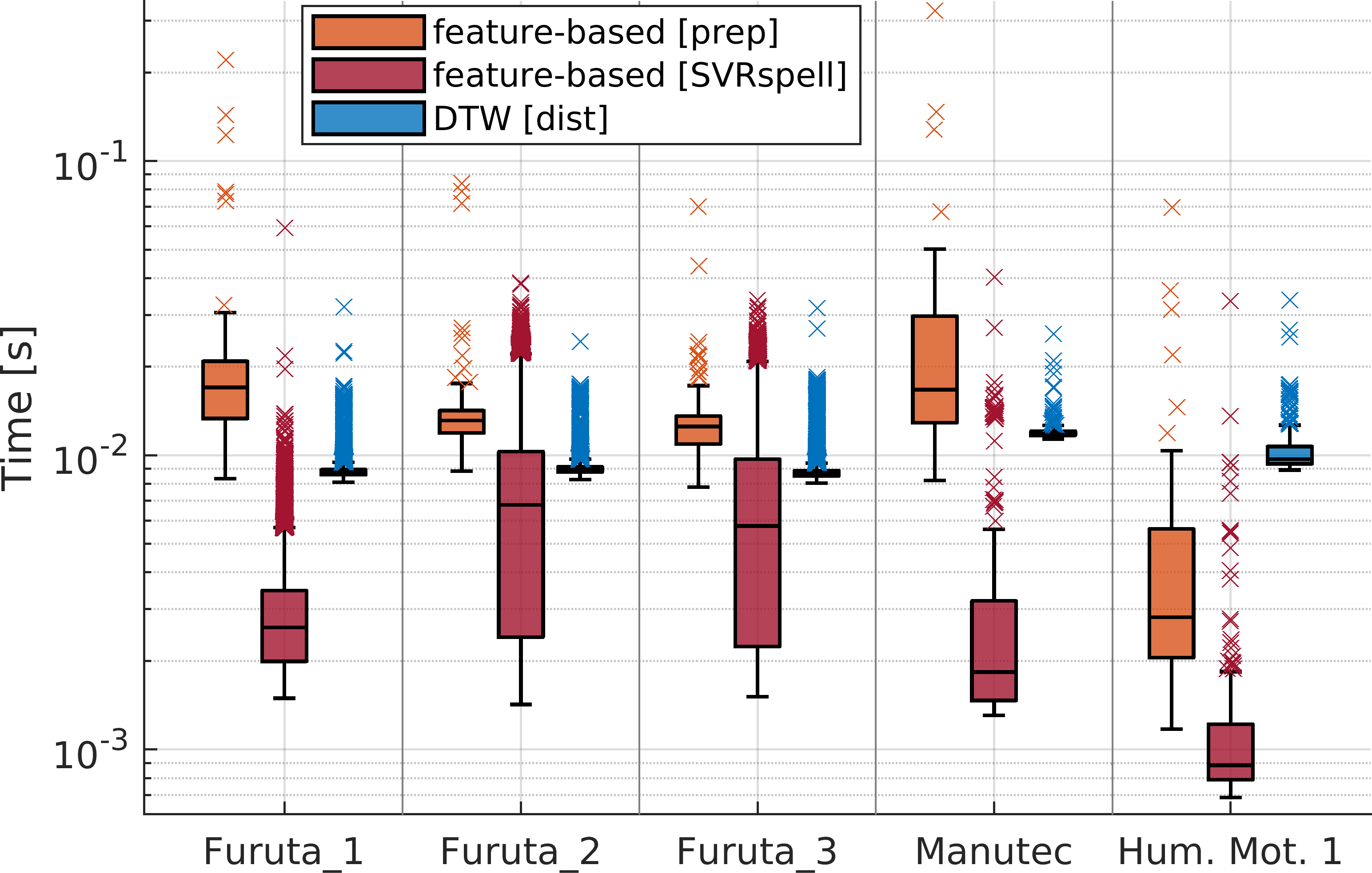}
    \caption{Computation time for a single distance computation between two trajectories on the test sets considered here. Computation time of the feature-based approach is splitted in precomputation step and \emph{SVRspell} distance computation.}
    \label{fig:evaluation_time}
\end{figure}

The runtime of the \emph{SVRspell}-based distance computation is important, as the number of calls to the distance computation is quadratic or even cubic in the number of trajectories for many clustering approaches \cite{murtagh_algorithms_2017}.
Fig. \ref{fig:evaluation_time} shows the time to extract the feature sequence (orange) as well as the times to compute the distance between two trajectories using DTW (blue) and feature-based approach (red). It must be noted that for this evaluation, the DTW and SVRspell implementations run as compiled code while the feature extraction is implemented in plain Matlab. 
The distance measure based on the feature sequence representation has clear advantages regarding computation time.

\section{Discussion, Conclusion and Outlook}\label{sec_conclusion}
The number of features has the largest impact on the runtime complexity of the new distance measure, which effectively decouples the computation of the distance from the time length of a trajectory. This is an advantage if long trajectories can be reduced to a small number of features. The number of features per trajectory depends on the trajectory itself, the feature classes used and the parameterization (thresholds).
An advantage of our method is flexibility in the selection of arbitrary many and arbitrary complex feature classes, which allows adaptation to specific problems and use cases. The parameters of the presented method are feature class dependent; typical parameters are thresholds below which features are considered as irrelevant and are discarded to keep the feature sequence as small as possible. It must be noted that our distance measure needs a sufficient number of distinct features in the trajectory graph to show its potential.

We have proposed a new approach to cluster trajectories that represent motion plans. It is based on a method \cite{elzinga_sequence_2003} to measure the distance between strings. We present an approach to compress trajectories to sequences of user-defined feature classes to make it applicable for trajectories.
Overall, the results show a reliable performance and several advantages compared to the proven DTW. In some cases, it outperforms this widely used distance measure.

In the field of human-robot interactions, further potential of the presented distance measure lies in the construction of hierarchical motion databases (as done, e.g., in \cite{yamane_synthesizing_2013, kulic_incremental_2012}), since the compressed feature-based representation of motions is memory efficient and allows fast comparisons.

\addtolength{\textheight}{-3cm}  

\bibliographystyle{IEEEtran}
\bibliography{IEEEabrv,biblio}

\end{document}